\pdfoutput=1

\documentclass[journal]{IEEEtran}
\usepackage{pgfplots}
  \pgfplotsset{compat=newest}
  
  \usetikzlibrary{plotmarks}
  \usetikzlibrary{arrows.meta}
  \usepgfplotslibrary{patchplots}
  \usepackage{grffile}
  \pgfplotsset{plot coordinates/math parser=false}
  \newlength\figureheight
  \newlength\figurewidth  
  
  \usepackage{cite}
  \usepackage{amsmath}
\usepackage{hhline}
\usepackage{verbatim }
\usepackage[hyperindex=true,colorlinks, citecolor=blue,menucolor=blue]{hyperref} %
\hypersetup{
	colorlinks=true, %
	citecolor=blue,
	filecolor=blue,
	linkcolor=blue,
	urlcolor=red,
	linktoc=all,     %
}

\usepackage{lineno}
\usepackage{amssymb, graphicx, amsmath,  psfrag, color, array, mathrsfs, setspace, algorithm, algpseudocode, mathtools, esint}
\DeclareMathOperator*{\argmax}{arg\,max}
\DeclareMathOperator*{\argmin}{arg\,min}
\usepackage{amsthm}

\newtheorem{theorem}{Theorem}

\allowdisplaybreaks

\newcommand{\lat}{{\theta}}

\newcommand{\expec}{\mathbb{E}}
\newcommand{\pol}{{\pi (a|s)}}

\newcommand{\pri}{{p_0 (a|s)}}

\newcommand{\bel}{{p(\lat|a,s)}}

\newcommand{\bayespost}{{p_0(\lat|a,s)}}

\newcommand{\trantheta}{{T_{\theta}(s'|a,s)}}

\newcommand{\rew}{R_{s,a}^{s'}}

\newcommand{\expectedUtilitystar}{{\expec_{\trantheta} \left[ \rew - \gamma F (s') \right]}}

\definecolor{amber}{rgb}{1.0, 0.49, 0.0}
\definecolor{americanrose}{rgb}{1.0, 0.01, 0.24}
\definecolor{amethyst}{rgb}{0.6, 0.4, 0.8}

\makeatletter
\let\MYcaption\@makecaption
\makeatother
\usepackage[font=footnotesize]{subcaption}
\makeatletter
\let\@makecaption\MYcaption
\makeatother

\begin{document}

\title{Reinforcement Learning with Subspaces \\ using Free Energy Paradigm}

\author{Milad~Ghorbani, %
         Reshad~Hosseini, %
		 Seyed~Pooya~Shariatpanahi,
        and Majid Nili~Ahmadabadi%
\thanks{The authors are with the School of ECE, College of Engineering, University of Tehran, Tehran, Iran (email: ghorbanimilad@ut.ac.ir; reshad.hosseini@ut.ac.ir; p.shariatpanahi@ut.ac.ir; mnili@ut.ac.ir).}%
}
\maketitle

\begin{abstract}
In large-scale problems, standard reinforcement learning algorithms suffer from slow learning speed. In this paper, we follow the framework of using subspaces to tackle this problem. We propose a free-energy minimization framework for selecting the subspaces and integrate the policy of the state-space into the subspaces. Our proposed free-energy minimization framework rests upon Thompson sampling policy and behavioral policy of subspaces and the state-space. It is therefore applicable to a variety of tasks, discrete or continuous state space, model-free and model-based tasks. Through a set of experiments, we show that this general framework highly improves the learning speed. We also provide a convergence proof.
\end{abstract}


\IEEEpeerreviewmaketitle

\section{Introduction}
Reinforcement learning (RL) is the most important paradigm for interactive learning where an agent learns through interaction with external environment~\cite{kaelbling1996reinforcement}. In recent years, RL has been studied in real-world tasks like robotics~\cite{abbeel2006learning,navarro2012real,imanberdiyev2016autonomous}, recommender system applications~\cite{lu2015recommender,da2016evolutionary}, and human-in-the-loop applications~\cite{holzinger2016interactive,zhang2017human,li2016dialogue}.

The main problem of RL is that at the beginning of the learning process, the agent has no knowledge and should try all possible actions in all different states. Large-scale RL problems with multidimensional state-space suffer from the curse of dimensionality~\cite{sutton1996generalization}. The main solution to confront this problem is generalization. Some generalization methods try to cluster similar states during learning and obtain an abstraction of the main problem and then solve the abstract problem.

Thompson sampling (see~\cite{thompson1933likelihood}) is a policy that efficiently balances between exploration and exploitation~\cite{gopalan2014thompson}. In~\cite{mandel2016efficient}, an efficient approach called TCRL, has been introduced that uses an idea similar to Thompson sampling for finding suitable clusters during the learning process.

In~\cite{abel2017near,abel2018state}, some approximation methods for state abstraction have been introduced. Unlike TCRL, these methods do not guarantee to converge to the optimal policy. However, they are more suitable for large-scale problems since they need less exploration.

Finding similar states for making abstraction also needs sufficient exploration and can suffer from the curse of dimensionality. Some other methods that partially solve this problem use a particular abstract RL problem called subspace, which is usually fixed in the beginning~\cite{firouzi2011interactive,daee2014reward,hashemzadeh2018exploiting}. A subspace is created by applying a transformation on the state space that reduces the number of features. Agent generalizes its experiences when it uses a subspace instead of the main space. In some states, this generalization may hinder the agent from having the optimal policy. This problem of subspace generalization is called perceptual aliasing (PA). Therefore, a concrete model that integrates policies from subspaces, and the main space was used in the literature (see ~\cite{firouzi2011interactive,daee2014reward,hashemzadeh2018exploiting}).
This concrete model often uses subspaces in early trials to improve the learning speed and then switches to the main space as the learning process goes on ~\cite{hashemzadeh2018exploiting}. This idea is also used in curriculum learning, where an agent learns in a sequence of subtasks before starting the main task. This sequence usually is generated automatically~\cite{silva2018object,schmidhuber2013powerplay} or designed by domain experts~\cite{narvekar2019learning}. The subtasks are often simpler versions of the main task and used to increase the speed of learning of the main task through transfer learning~\cite{narvekar2020curriculum}. Recently, curriculum learning methods have shown promising results in solving complex games such as Backgammon~\cite{tesauro1995temporal}, Go~\cite{silver2016mastering}, and Starcraft~\cite{vezhnevets2016strategic}. One of the shortcomings of these methods is that making an appropriate sequence of subtasks itself needs a significant amount of time~\cite{narvekar2020curriculum}. Therefore, the objective of this paper is to determine the suitable subspace to learn before learning the main task without using any prior information about different spaces.

In this paper, we investigate the problem of selecting suitable space and also integrating the policies of subspaces and the main space, and develop a method based on free energy minimization.
Free energy model is used to model the behavior of decision-maker when it has some information constraints~\cite{ortega2011information,ortega2013thermodynamics}. Negative free energy can also be seen as a criterion for the reliability of a decision-making model in a particular state~\cite{stephan2009bayesian}. Subspaces have an inherent information limit which is introduced as PA. 
Therefore, we propose a method in which 
the agent creates models in the main space and some selected subspaces and then estimates the uncertainty of these models. Then, it uses free energy model to amalgamate these informations and obtain a good policy to interact with the environment. The free energy model is used to tackle with natural PA of subspaces and also limited accuracy of uncertainty estimations which is inevitable and can cause considerable inefficiencies.

Our proposed framework for subspace generalization is very general. Previous works by other members of our group exploited the idea of subspaces in model-free~\cite{firouzi2011interactive,daee2014reward} or model-based~\cite{hashemzadeh2018exploiting} settings. In contrast, our method is implementable in different model-free, model-based, and deep RL settings.
 Through several experiments, we compare the performance of our method with other subspace-based methods. Our method significantly has a better performance
than previous subspace-based methods.

\section{Problem Statement and Assumptions}
The agent lives in an unknown Markov decision process (MDP) with discrete action space defined by a tuple $\{S,A(.),T(.| ., .),R(., .),\gamma\}$, where $S$ is the state space which is discrete or continuous, $A(s)$ is the action space in the state $s$, $T(s'|s,a)$ is the transition probability from state $s$ to state $s'$ by performing action $a$, $R(s,a)$ is the expected reward for state-action $(s,a)$, and $0\le\gamma\le1$ is a discount factor for the accumulated reward~\cite{bellman1957markovian}.

Policy is defined as a mapping of state $s$ to action $a$ denoted by $\pi(a|s)$. The agent's goal is to find the optimal policy $\pi^*(a|s)$, which is a policy that maximizes the discounted return which is defined as follows~\cite{sutton2018reinforcement}
\begin{equation}
G = r_{1}+\gamma r_{2} + \cdots =\sum_{k=0}^{\infty}\gamma^k r_{k+1},
\end{equation}
where $k$ is time-step, $r_{k+1}$ is the instant reward.

\section{Preliminaries}

In this section, we first present Thompson sampling for policy estimation in reinforcement learning. We give an overview of interval estimation methods and dropout that are used for computing policy using Thompson sampling. Afterward, we give an overview of using free energy paradigm in decision making, and its connection to Thompson sampling.

\subsection{Thompson Sampling Policy Estimation}\label{ss:Thompson_estimation}
Thompson sampling selects each action in a state according to the probability of the optimality of that action~\cite{ortega2014generalized}. This probability can be computed using belief distributions, that is the probability distribution of values under model uncertainty, for actions in each state. In reinforcement learning framework, belief distributions are updated during the learning process. The policy of Thompson sampling is given by:
\begin{multline}\label{eq:Thompson_Policy_Calc}
  \pi_{TS}(a_i|s) = P\bigg(\bigcap_{j\neq i}\Big\{Q(s,a_i)>Q(s,a_j)\Big\}\bigg) =  \\
 \int_{-\infty}^{\infty} p_Q(x_i|s,a_i) \prod_{j\neq i} P(x_i>Q(s,a_j)) dx_i=\\
 \int_{-\infty}^{\infty} p_Q(x_i|s,a_i) \prod_{j\neq i}\int_{-\infty}^{x_i} p_Q(x_j|s,a_j) dx_j dx_i,
\end{multline}
where $Q(s,a_i)$ is the value of action $a_i$ in state $s$, and $p_Q(.|s,a_i)$ is the belief distribution for action $a_i$ in state $s$. For computing the policy using Thompson sampling, we need to compute belief distributions. This can be computed by direct application of Bayes rule, but it is not usaually used due to its computational complexity~\cite{gal2016dropout}. This problem is tackled commonly by approximate belief distribution. One way to approximate belief distribution is by computing confidence intervals~\cite{jaksch2010near}. In the following, we review different methods for estimating belief distrubution. Assuming belief distribution to be Gaussian, its mean and variance can be obtained from confidence intervals. Dropout is another method to approximate belief distribution in continuous environments~\cite{gal2016dropout}.

\subsubsection{Model Free Interval Estimation}
The actual value of doing action $a$ in state $s$ is denoted as $Q(s,a)$, and its estimated value (Q-value) as $\hat{Q}(s,a)$. Model free Q-values are updated by the following rule in Q-learning:
\begin{multline}
\hat{Q}_{t+1}(s_{t+1},a_{t+1}) = \hat{Q}_t(s_t,a_t)+\eta_t(s_t,a_t)[r_t(s_t,a_t)+\\
\gamma \max_{a}\hat{Q}_t(s_{t+1},a) -\hat{Q}_t(s_t,a_t)],
\end{multline}
where $\eta_t(s,a)$ is learning rate, $r_t(s,a)$ is immediate reward, and $t$ is time-step. A confidence interval on $Q(s,a)$ can be obtained by using the following bound~\cite{daee2014reward}, that is valid when sample size $n(a,s)$ is moderately large or when the reward distribution is Gaussian~\cite{daee2014reward}
\begin{equation}\label{eq:Ped_Bound_Inequality}
P\left( \hat{Q}(s,a)-\mu< Q(s,a) <\hat{Q}(s,a)+\mu\right)=1-\nu,
\end{equation}
\begin{equation}\label{eq:Ped_Bound}
\mu = t_{\frac{\nu}{2},n(s,a)-1}\times \frac{\overline{std}(s,a)}{\sqrt{n(s,a)}},
\end{equation}
where $t_{\frac{\nu}{2},n(s,a)-1}$ is the one-side t-value with $\frac{\nu}{2}$ confidence level and $n(s,a)-1$ degrees of freedom, $n(s,a)$ is the sample size for action $a$ in state $s$, and $\overline{std}(s,a)$ is the estimated standard deviation of the underlying reward distribution defined by:
\begin{equation}
\overline{std}(s,a)=\sqrt{\frac{n(s,a)\sum_t \tilde{Q}_t(s,a)^2-(\sum_t \tilde{Q}_t(s,a))^2}{n(s,a) \times (n(s,a)-1)}},
\end{equation}
where $\tilde{Q}_t$ is a monte-carlo sample of $Q$-values and is defined by:

\begin{equation}
\tilde{Q}_t(s,a) = \sum_{k=k_s}^{k_T} {\gamma ^{k-k_s} r_k},
\end{equation}
where $k_s$ is the time-step in an episode that agent visits state $s$, $k_T$ is the time-step that the agent visits the terminal state, and $r_k$ is the reward that is taken by the agent in the time-step $k$.
 The bound introduced in \eqref{eq:Ped_Bound_Inequality} has been used and works reasonably well in many practical experiments~\cite{daee2014reward,daee2014developmental}.

\subsubsection{Model Based Interval Estimation}
Assume the agent is in state $s$, performing action $a$ and goes to state $s'$ and receives immediate reward $r_t(s,a)$, and the sample size for this transition is denoted by $n(s,a,s')$. Then, the model parameters such as approximated transition probability $\hat{T}(s'|s,a)$ and the expected reward $\hat{R}(s,a)$ are updated by:
\begin{equation}
\hat{T}(s'|s,a)=\frac{n(s,a,s')+1}{n(s,a)+1},
\end{equation}
\begin{equation}
\hat{R}(s,a)=\frac{\hat{R}(s,a)\times n(s,a,s')+r_t(s,a)}{n(s,a)+1}.
\end{equation}

The Q-values are estimated by solving the following Bellman equation\cite{bellman1957markovian}:
\begin{equation}
\hat{Q}_\pi (s,a)=\hat{R}(s,a)+\sum_{s'} \hat{T}(s'|s,a)\left(\gamma \max_{a'} \hat{Q}_\pi (s',a')\right).
\end{equation}

To obtain confidence intervals for Q-values, confidence intervals of model parameters such as $\hat{R}(.)$ and $\hat{T}(.)$ are needed. We can use Hoeffding 
\cite{hoeffding2014probability} and Weissman
\cite{weissman2003inequalities} inequalities to obtain these confidence intervals, respectively.
  
The lower and upper bounds are estimated using UCRL2 algorithm\cite{jaksch2010near,hashemzadeh2018exploiting}.
 The algorithm solves the following equations to obtain lower $Q_l$ and upper $Q_u$ bounds on Q-values:
\begin{equation}
\begin{aligned}
 & \hat{Q}_u(s,a) = \left( \hat{R}(s,a)+\epsilon_R \right) \\
 & + \max_{\tilde{T}(s'|s,a)\in CI_T} \sum_{s'} \tilde{T}(s'|s,a)\left( \gamma \max_{a'}\hat{Q}_u(s',a') \right),
\end{aligned}
\end{equation}
\begin{equation}
\begin{aligned}
 & \hat{Q}_l(s,a) = \left( \hat{R}(s,a)-\epsilon_R \right) \\
 & + \min_{\tilde{T}(s'|s,a)\in CI_T} \sum_{s'} \tilde{T}(s'|s,a)\left( \gamma \max_{a'}\hat{Q}_l(s',a') \right).
\end{aligned}
\end{equation}

In these equations $\epsilon_R$ is half of confidence interval for estimated reward $\hat{R}(s,a)$, and $CI_T$ is the confidence set for the estimated transition probability $\hat{T}(s'|s,a)$. 

\subsubsection{Dropout for Bayesian Approximation}\label{ss:Dropout for Bayesian Approximation}
In continuous domains, one possible approach to calculate Thompson sampling policy is also confidence interval estimation as explained in 
\cite{white2010interval}. However, the method of \cite{white2010interval} can only be applied for specific continuous domains and it is not straightforward to generalize the idea for the case of neural networks like Q-network where a neural network is used for estimating Q-values of actions. We can obtain a sample of the belief distribution of Q-values by applying dropout before each weighting layer of the Q-network and computing its output~\cite{gal2016dropout}. An agent who uses Thompson sampling selects the action that has maximum value among all actions for all networks created by dropout. If the total number of networks created by dropout $N$ is large enough, Thompson sampling policy for action $a_i$ would be fairly accurate and is given by:
\begin{equation} \label{eq:Dropout_TS}
\pi_{TS}(a_i|s) = \frac{n_s(a_i,s)}{N},
\end{equation}
where $n_s(a_i,s)$ is number of times that action $a_i$ is selected by the network (has the maximum value among all action) with dropout rate of
 $p$. The advantage of this method is that it obviates the requirement for interval estimation for calculating Thompson sampling policy. 

\subsection{Free Energy Model of Decision Making}\label{ss:free_energy_model_of_DM}
Free energy model is used in the literature to model decision-making under information constraints (see~\cite{ortega2013thermodynamics}). The optimization problem for finding the optimal policy in this context can be given by
\begin{equation*}
\pi^*(a|s) = \argmin_{\pi(a|s)} F(s;\pi(a|s)),
\end{equation*}
\begin{equation}\label{eq:free_energy_model_of_decision_making}
F(s;\pi(a|s)) = \expec_{\pi(a|s)}\left[\frac{1}{\alpha}\log \frac{\pol}{\pri} - U(a,s)\right],
\end{equation}
where $F(s)$ is the free energy of an agent in a particular state $s$ when it uses the policy $\pol$ and has a reference prior policy $\pri$. In~\eqref{eq:free_energy_model_of_decision_making}, $U(a,s)$ is the utility of action $a$ in state $s$. The inverse temperature $\alpha$ is the trade-off constant between the information cost and the expected utility. 

This model is often used to model the behavior of an agent for decision making under limited computational resources, e.g. when the agent has to choose an action between too many actions and actually can try a subset of them in a limited time~\cite{tishby2011information,ortega2011information,gottwald2019bounded,trujillo2019mental}, or when the agent can use multiple experts and needs to choose one of them quickly even if the selected one is sub-optimal~\cite{hihn2019hierarchical,hihn2019information}.

The model of~\eqref{eq:free_energy_model_of_decision_making} does not consider model uncertainty during information processing, which is necessary to better exploration-exploitation balance. In order to model uncertainty, one should consider a two-step version of~\eqref{eq:free_energy_model_of_decision_making}. A latent variable $\theta$ is introduced which determines the environment model $T_{\theta}(s'|a,s)$. The agent has a two-step minimization on the following nested free energy functional~\cite{ortega2014generalized}
\begin{multline}
\pi^*(a|s) = \argmin_{\pi(a|s)}\min_{\bel} F(s;\pi(a|s),\bel),\\
F(s;\pi(a|s),\bel) = \expec_{\pol} \bigg[ \frac{1}{\alpha} \log{\frac{\pol}{\pri}}+\\
\expec_{\bel}\left[\frac{1}{\beta}\log{\frac{\bel}{\bayespost}}-U(a,\lat,s)\right]\bigg],\label{eq:two_step_free_energy}
\end{multline}
where $p_0(\theta|a,s)$ is the prior belief on the latent variable, $\bel$ is the biased belief and $\beta$ is a hyperparameter that indicates how much the agent trusts the prior belief. One can consider the latent variable $\theta$ as a vector to be the parameters of the model, and $p(\theta|a,s)$ as a distribution in the space of parameters. When the space of parameters increrases, solving the two-step optimization problem in~\eqref{eq:two_step_free_energy} may become intractable.

In~\cite{grau2016planning}, this free energy model is used to solve an MDP problem under information constraints using a free energy iteration method. The free energy of an MDP problem is a special case of~\eqref{eq:two_step_free_energy} where
\begin{equation}\label{eq:generalized_bellman}
	U(a,\lat,s) = \expectedUtilitystar.
\end{equation}

If we use \eqref{eq:generalized_bellman} in \eqref{eq:two_step_free_energy}, we get a generalization to Bellman equation for value maximization when we have model uncertainties and also information constraints.

Decision-making in~\eqref{eq:two_step_free_energy} can be seen as a model for transforming the prior information under policy $p_0$ into posterior information $p$ by taking into account the utility gains (or losses), and the transformation costs arising from information processing and model uncertainties. It can be proven that for single-state decision-making problems, Thompson sampling policy is decision making under this model for the particular case of parameters $\alpha$ and $\beta$~\cite{ortega2014generalized}.

\section{Proposed Method}
In this paper, we assume the task is a Markov decision process (MDP) where the agent wants to both learn the parameters of the task and optimal policy using the reinforcement learning framework. Apart from the original space which is MDP, the agent learns the parameters of the subspaces and uses the subspaces for decision making. A subspace is a sub-dimensional space of the original space, created by applying an arbitrary transformation on the state-action space and modifies the space of state-action space as follows
\begin{equation}\label{eq:subspace_in_original} 
	s_{sub} = \Phi(s),
\end{equation}
where $\Phi$ is an arbitrary transformation that maps state $s$ to state $s_{sub}$ with lower number of features. This transformation might be only feature selection. For example in a 2D maze environment, a subspace can be x-position in each state. Since subspaces are smaller and have fewer parameters, the agent learns faster in them which causes better generalization and expediting the learning process. But, we need to find out which subspace or the original space should be chosen in each state for getting the best generalization during the learning process.

\begin{figure}[!t] 
\centering
\includegraphics[width=3.4in]{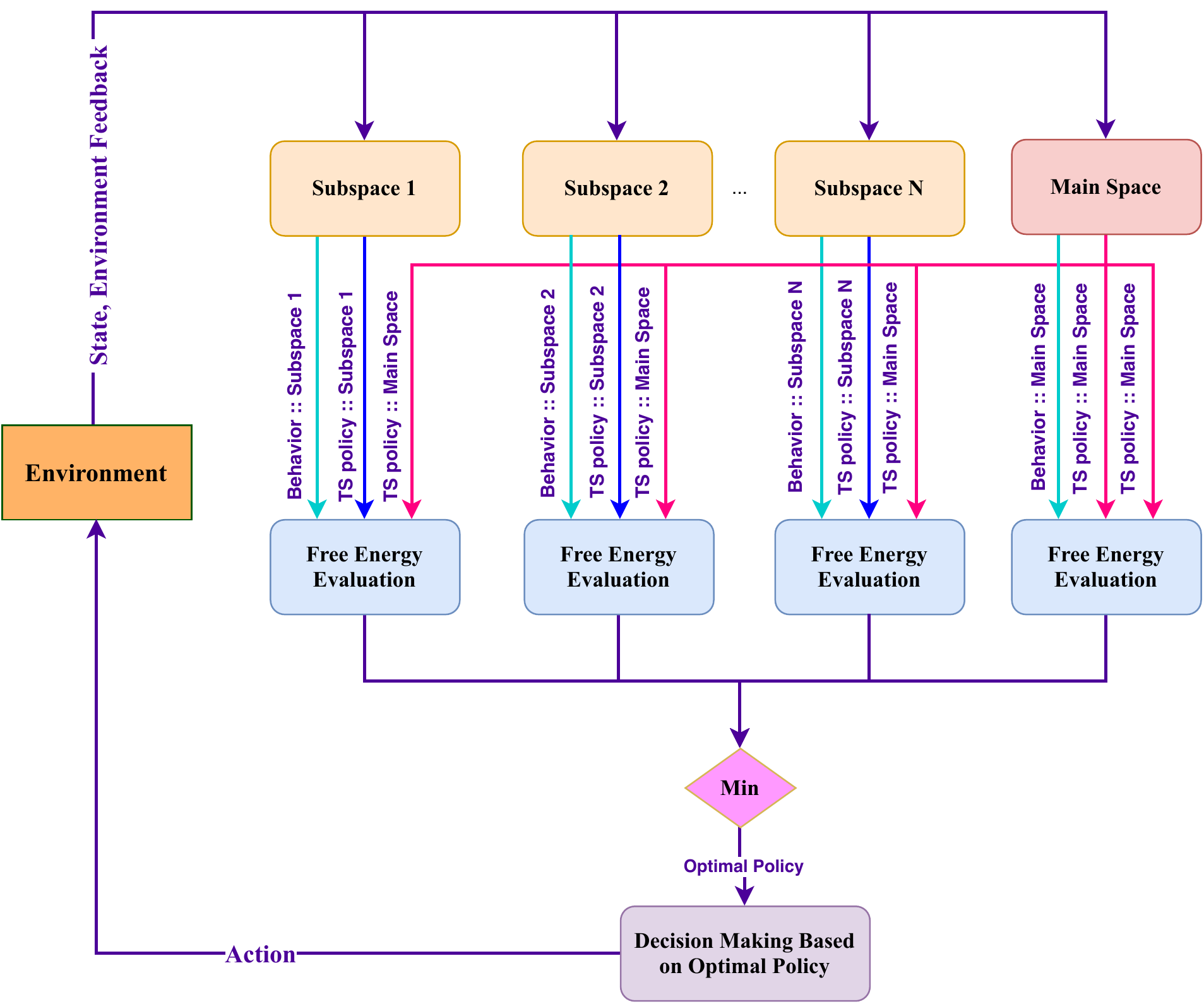}
\caption{ Information flow of the proposed method. The agent finds the optimal policy that minimizes free energy and evaluate the optimal free energy for subspaces and the main space, then it uses the optimal policy of the space with minimum free energy to interact with the environment.}
\label{fig:Method_Chart}
\end{figure}

In this work, the perspective to find the most appropriate subspace is minimizing free energy. In the free energy formulation, the utility in each space is a function of Thompson sampling, and the regularizing term encourages the policy of the space to be close to an arbitrary behavioral policy and also remain consistent with the main space utility to avoid perceptual aliasing. Then, it compares the free energy to select the original space or the most appropriate subspace and uses the computed policy from free energy minimization to interact with the environment. Fig. \ref{fig:Method_Chart} depicts the information flow in our proposed method which we call free energy with Thompson sampling utility, abbreviated with FETS in the rest of the paper.

\subsection{Selection with Subspaces using Free Energy Model}
An agent has several task models (subspaces or the main space), that are updated using the feedback from the external environment. For the model $m$ of each space, we use ~\eqref{eq:free_energy_model_of_decision_making} to calculate the free energy of the model, and we define the utility $U(a,s,m)$ to be negative informational surprise about optimality of action $a$ in state $s$ in that model. Thompson sampling policy is equal to optimality of actions (see ~\eqref{eq:Thompson_Policy_Calc}), so we define the utility as
\begin{equation}\label{eq:utility} 
	U(a,s,m) = \log{\pi_{TS}(a|s,m)}.
\end{equation}

This utility function (which is a function of Thompson sampling policy) handles exploration-exploitation trade-off, and has some computational benefits, because one of the optimizations in~\eqref{eq:two_step_free_energy} which is intractable when the model parameter space expands, is substituted by the estimation of Thompson sampling policy.
To deal with the PA problem of subspaces, we use the following constraint that forces the agent to consider the main space utility in each space
\begin{equation}\label{eq:constraint_1}
	\expec_{\pi(a|s,m)} [ U(a,s,m)] - \expec_{\pi(a|s,m)}[ U(a,s,m_{Main})]<K_1,
\end{equation}
where $m$ is a model that can be subspaces or the main space, and $m_{Main}$ is the main space\footnote{In the case $m=m_{Main}, $~\eqref{eq:constraint_1} always holds because we do not have PA in the main space.}. The constraint~\eqref{eq:constraint_1} ensures that expected utility of the main space is not much less than the expected utility of the subspace. We also want the policy to be close to an arbitrary behavioral policy, so we consider the following constraint
\begin{equation}\label{eq:constraint_2}
	\mathbb{D}_{KL}( \pi(a|s,m) || \pi_B(a|s,m)) <K_2,
\end{equation}
where $\pi_{B}(a|s,m)$ is an arbitrary behavioral policy.
The constraint~\eqref{eq:constraint_2} is very useful when we have nonaccurate estimations, because we can limit the effect of low accuracy estimations of uncertainties to the policy of the agent by selecting a behavioral policy like $\epsilon$-greedy that does not use the uncertainty estimations.
By considering the utility function of~\eqref{eq:utility} and constraints~\eqref{eq:constraint_1} and~\eqref{eq:constraint_2}, this problem can be cast as following free energy minimization problem 
 \renewcommand{\algorithmicrequire}{\textbf{Input:}} 
  \begin{algorithm}[!t] 
  \newcommand{\NewComment}[1]{ {\hfill$//$ #1}} %
   \caption{Free Energy Evaluation}\label{alg:al_free_energy}
   \selectfont
  	    \footnotesize {\textbf{Function} FreeEnergy($ \pi_{TS}(a|s,m), \pi_{TS}(a|s,m_{Main}),\pi_{B}(a|s,m),\alpha,\beta $)}
		
    \begin{algorithmic}[1] 
      
        \Require 
 	    {$\pi_{TS}(a|s,m)$ is Thompson sampling policy for model $m$, $\pi_{TS}(a|s,m_{Main})$ is Thompson sampling policy for the main space, $\pi_{B}(a|s,m)$ is an arbitrary behavioral policy, $\alpha$ is the parameter to be restricted to the behavioral policy and $\beta$ is the the generalization parameter of the agent} 
		
		\State $U(a,s,m) = \log{\pi_{TS}(a|s,m)}$ \NewComment{Utility definition~\eqref{eq:utility}}
		\State $\tilde{U}(a,s,m) = U(a,s,m) -  \frac{1}{\beta} (U(a,s,m) - U(a,s,m_{Main}))$  
		\State $Z(s,m) = \sum_{a}\pi_B(a|s,m) e^{\alpha \tilde{U}(a,s,m)}$
		\State $\pi^*(a|s,m) = \frac{1}{Z(s,m)}\pi_B(a|s,m) e^{\alpha \tilde{U}(a,s,m)}$ \NewComment{Using~\eqref{eq:OptimalPolicy}}
		\State $F(s,m) = \sum\limits_{a}\pi^*(a|s,m)\left[ \frac{1}{\alpha} \log \frac{\pi^*(a|s,m)}{\pi_B(a|s,m)} - \tilde{U}(a,s,m)\right]\newline$\NewComment{Using \eqref{eq:optimization_subspaces}}\\
		
		\Return {$ F(s,m) , \pi^*(a|s,m)$}

\end{algorithmic}
\end{algorithm}
\begin{equation*}
\pi^*(a|s,m) = \argmin_{\pi(a|s,m)} F(s,m,\pi(a|s,m)),
\end{equation*}
\begin{multline}
	F(s,m,\pi(a|s,m)) = 
	\expec_{\pi(a|s,m)}\bigg[\frac{1}{\alpha}\log \frac{\pi(a|s,m)}{\pi_{B}(a|s,m)}+ \\
	           \frac{1}{\beta}\log{\frac{\pi_{TS}(a|s,m)}{\pi_{TS}(a|s,m_{Main})}} - \log{\pi_{TS}(a|s,m)}\bigg],\label{eq:optimization_subspaces}
\end{multline}
where parameters $\beta$ and $\alpha$ are the Lagrange multipliers for the constraints~\eqref{eq:constraint_1} and~\eqref{eq:constraint_2}, and they have a direct relationship with $K_1$ and $K_2$, respectively.
 Parameters $\beta$ and $\alpha$ control the force of making the agent behavior in the state to be similar to that of the main space and given behavioral policy, respectively. 
 The optimization problem of~\eqref{eq:optimization_subspaces} is identical to the free energy framework in ~\eqref{eq:free_energy_model_of_decision_making} and the corresponding utility function is
 \begin{equation}
	\tilde{U}(a,s,m) = -\frac{1}{\beta}\log{\frac{\pi_{TS}(a|s,m)}{\pi_{TS}(a|s,m_{Main})}}+U(a,s,m),
\end{equation}
so the optimal solution for this problem is~\cite{ortega2013thermodynamics}
\begin{equation*}
	\pi^*(a|s,m) = \frac{1}{Z(s,m)}\pi_{B}(a|s,m) e^{\alpha \tilde{U}(a,s,m)},
\end{equation*}

\begin{equation}\label{eq:OptimalPolicy}
	Z(s,m) = \sum_a \pi_{B}(a|s,m) e^{\alpha \tilde{U}(a,s,m)}.
\end{equation}

Finally, the best space is given by
\begin{equation}
	m^* = \argmin_{m}F(s,m,\pi^*(a|s,m)).
\end{equation}

Then, the agent samples an action according to the policy of $\pi^*(a|s,m^*)$ to ineract with its environment. After that, it updates all models in parallel due to the selected action and the feedback of the environment. Algorithm \ref{alg:al1} illustrates our framework for discrete and continuous domains. Next, in Theorem \ref{th:1}, we investigate the conditions for convergence of algorithm in discrete domain.

\begin{theorem}\label{th:1}
Let $\mathbb{M}$ be an MDP with finite state space $\mathbb{S}$,
 and finite action space $\mathbb{A}$. In the proposed scheme, if an agent uses finite $\alpha>0$ and $\beta>1$ and uses $\epsilon$-greedy as its behavioral policy, then the agent finally uses the main space policy. Therefore, it guarantees convergence to the optimal policy\footnote{In practice, the utility function in~\eqref{eq:utility} might get numerally not defined values as Thompson sampling policy goes to zero. Therefore, we add a small constant to the probabilities to avoid these cases. }.
\end{theorem}
The detailed proof for Theorem~\ref{th:1} is given in Appendix~\ref{apx:proof_th1}.

\renewcommand{\algorithmicrequire}{\textbf{Input:}} 
  \begin{algorithm} [t]
  \newcommand{\NewComment}[1]{ {\hfill$//$ #1}} %
   \caption{Concurrent RL using subspaces with FETS}\label{alg:al1}
   \selectfont
  	    \footnotesize {\textbf{Function} FETS($ s, r, I, D, W , \pi_B, p,\alpha,\beta$)}
		
    \begin{algorithmic}[1] 
      
        \Require 
 	    {$s$ is the state observed from the environment, $r$ is the reward from environment, $I$ consists of values, Q-networks and confidence intervals, $D$ is the domain type (Discrete or Continuous), W is a set consisting of subspaces and the main space models. Model can be Q-table, MDP, neural network and etc, $\pi_B$ is an arbitrary behavioral policy, $p$ is dropout probability, $\alpha$ is the agent parameter to remain restricted to the behavioral policy, $\beta$ is the agent generalization characteristic}\\ 
 	    \footnotesize {\textbf{Initialize}} $F(s,m)$\\
 	    \footnotesize {\textbf{Update}} $I$
        \ForAll{$m \in W$} 
        
        	\If{$D$ is discrete (Model-free or Model-based)}
        	\ForAll{$a_i \in A$}
        		\State Use $I.CI$($s$) to form approximated distribution $P_i$
        		\State Use \eqref{eq:Thompson_Policy_Calc} to compute $\pi_{TS}$ 
        	\EndFor
        	\EndIf
        	\If{$D$ is continuous}
        		\State $n_s(.) = 0$
        		\For{$i = 1$ to $N$}
        			\State $Q = I.NN.forward(dropout = p)$ \NewComment{A forward pass in the Q-network with dropout rate of $p$}
        			\State $a_b = \argmax{Q} $
           			\State $n_s(a_b) = n_s(a_b) + 1$
        		\EndFor
        		\State Compute $\pi_{TS}$ using \eqref{eq:Dropout_TS}
        	\EndIf
	    \State $F(s,m) , \pi^*(a|s,m) = $FreeEnergy$(\pi_{TS}(a|s,m),\pi_{TS}(a|s,m_{Main}),\newline \pi_B(a|s,m),\alpha,\beta)$  \NewComment{Use Algorithm \ref{alg:al_free_energy}. $m_{Main}$ is the main space.}
	    \EndFor
        \State $m^* = \argmin{F(s,m)}$  
        \State Sample according to $\pi^*(a|s,m^*)$ to get action $a$\\
		\Return {$a$}

\end{algorithmic}
\end{algorithm}

\section{Experimental Results}
We test our framework in both discrete and continuous domains and for model-free and model-based reinforcement learning\footnote{The codes for reproducing the results of this section can be downloaded from \url{https://www.dropbox.com/sh/kz8q1685lmaxl6m/AAAMHGwaOmHiWOuKn-grd-rma?dl=0}.}. 
\subsection{Discrete Domain}
\begin{figure}[!t] 
\centering
\includegraphics[width=0.5\textwidth]{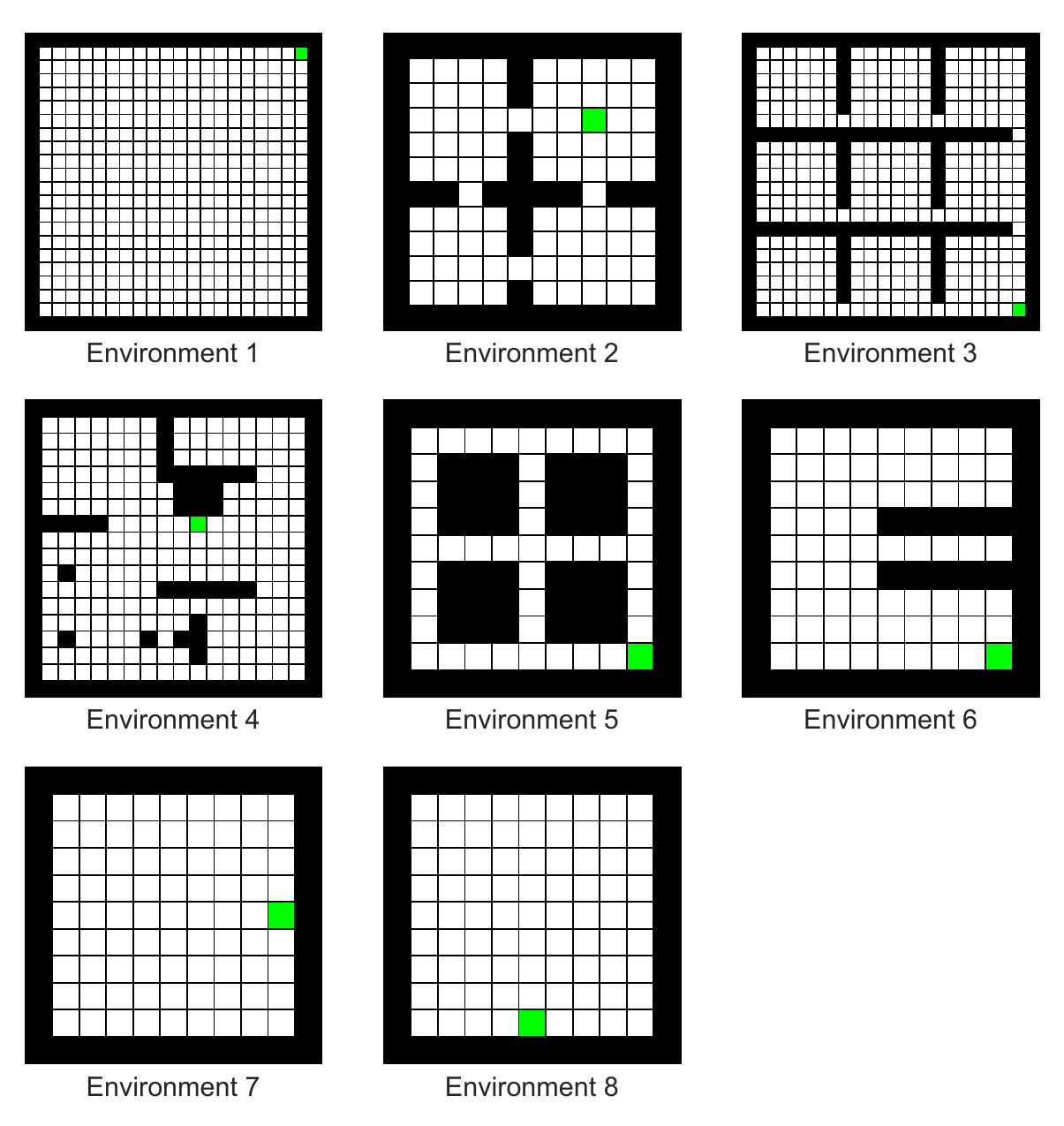}
\caption{2-D maze environments.}
\label{fig:2DEnvironments}
\end{figure}
In the discrete case, we compare our framework (FETS) with other discrete off-policy approaches using $\epsilon$-greedy policy as the behavioral policy. We select the following methods based on some key feaures: subspaces usage, space selection criteria, and the kind of policy.
\begin{enumerate}
\item MB: Our model-based approach ignoring subspaces. The policy of the agent is the behavioral policy. 
\item MB-FE: Our model-based approach ignoring subspaces. The policy of the agent is given in~\eqref{eq:OptimalPolicy} instead of using the behavioral policy. 
\item MB-CDM: Model-based approach with generalization in subspaces termed MoBLeS that integrates policies of subspaces using a confidence degree model (CDM)\cite{hashemzadeh2018exploiting}. 
\item MB-FETS-B: Model-based FETS method where the policy is the behavior policy in the selected space.

\item MB-FETS-FE: Model-based FETS method where the policy is given in~\eqref{eq:OptimalPolicy} for the selected space. 
\item MF: Famous $Q(\lambda)$ learning method~\cite{watkins1989learning} by selecting the
best decay rate of the eligibility trace ($\lambda$).

\item MF-FE: $Q(\lambda)$ method where the agent uses the policy that is given in~\eqref{eq:OptimalPolicy}. 

\item MF-LUS: $Q(\lambda)$ with generalization in subspaces that integrates policies of subspaces with the least uncertain source (LUS) method\cite{daee2014reward}. 
\item MF-FETS-B: Model-free FETS method where the policy is the behavior policy in the selected space.
\item MF-FETS-FE: Model-free FETS method where the policy is given in~\eqref{eq:OptimalPolicy} for the selected space. 
\end{enumerate}
We test our framework on 2-D maze environments and a  real-time strategy game called StarCraft:Broodwar (SC:BW)\cite{wender2012applying}. In these experiments, the parameters of the agent integration and generalization are $\alpha = 4$ and $\beta = 7$, respectively. However, the algorithm is not very sensitive to the exact values of these parameters, and the results are similar in a wide interval around these parameters.
   
\subsubsection{2-D Maze Environment}
We simulate 2-D maze environments introduced in\cite{hashemzadeh2018exploiting} and 4 other new environments (Fig. \ref{fig:2DEnvironments}). In these environments, the agent can move in four directions (up, right, down and left). The task is not deterministic, and each action has its real effect with a probability of 0.9; otherwise, the agent moves in a random direction. The agent is given reward reaching the goal (the green cell of the environments) and is punished by hitting the barriers. There is a constant cost of performing each action. The state is represented by location X and Y (the vertical and horizontal axis). More information on the setting of 2-D maze environment can be found in Appendix~\ref{apx:2d_maze}.

In 2-D maze environments, two primitive subspaces are created by keeping either X or Y features. Some environments are generalizable in these subspaces (for example environment 1,5,7 and 8) and some environments are designed to be nongeneralizable (for example environment 4 and 6).
\begin{figure}[!t] 
\centering
\includegraphics[width=0.5\textwidth]{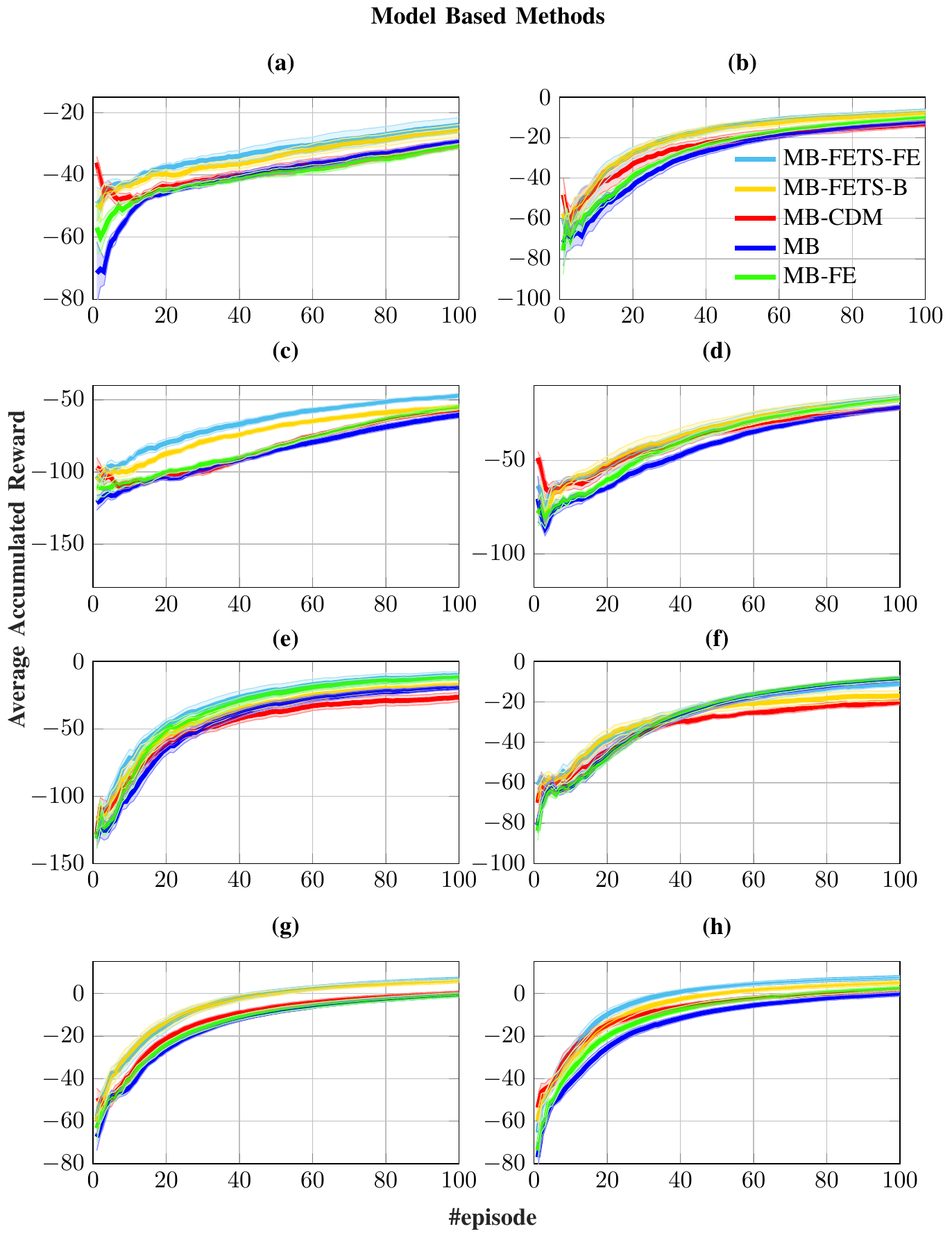}
\caption{(a)-(h) Average accumulative reward simulating model-based approaches for environments 1-8 in 15 runs.}
\label{fig:AvRewMB}
\end{figure}

\begin{figure}[!t] 
\centering
\includegraphics[width=0.5\textwidth]{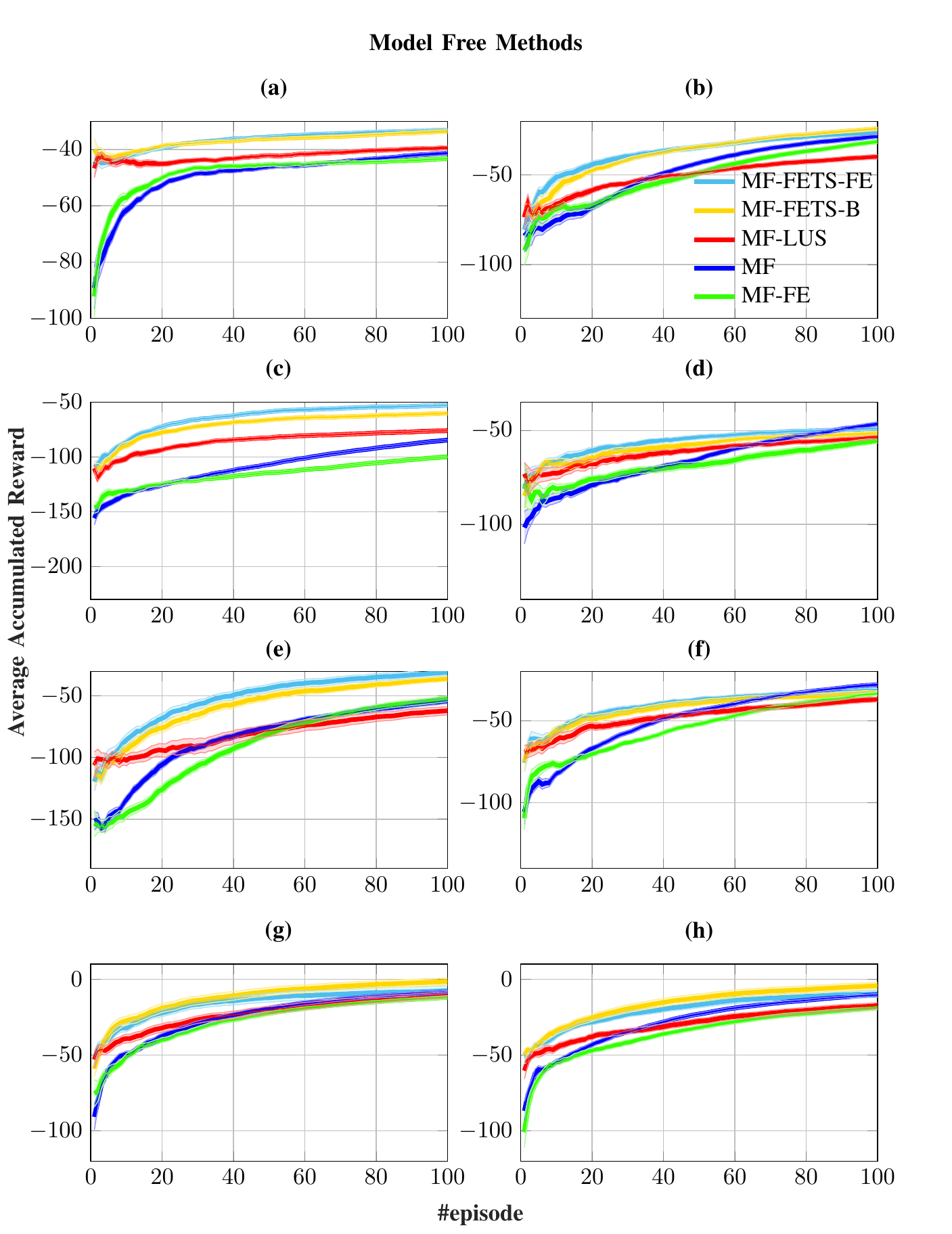}
\caption{(a)-(h) Average accumulative reward simulating model-free approaches for environments 1-8 in 15 runs.}
\label{fig:AvRewMF}
\end{figure}

In our simulations at the beginning of each episode, the starting state is randomized between possible starting states (empty non-goal cells). We record average accumulated reward for each method and average free energy over all states for subspaces and the
main space.
The average accumulated reward for model-based and model-free approaches are depicted in Fig. \ref{fig:AvRewMB} and Fig. \ref{fig:AvRewMF}, respectively. Our framework has a significantly better average accumulated reward in both model-based and model-free RL compared to the other methods.

\begin{figure}[!t] 
\centering
\includegraphics[width=0.5\textwidth]{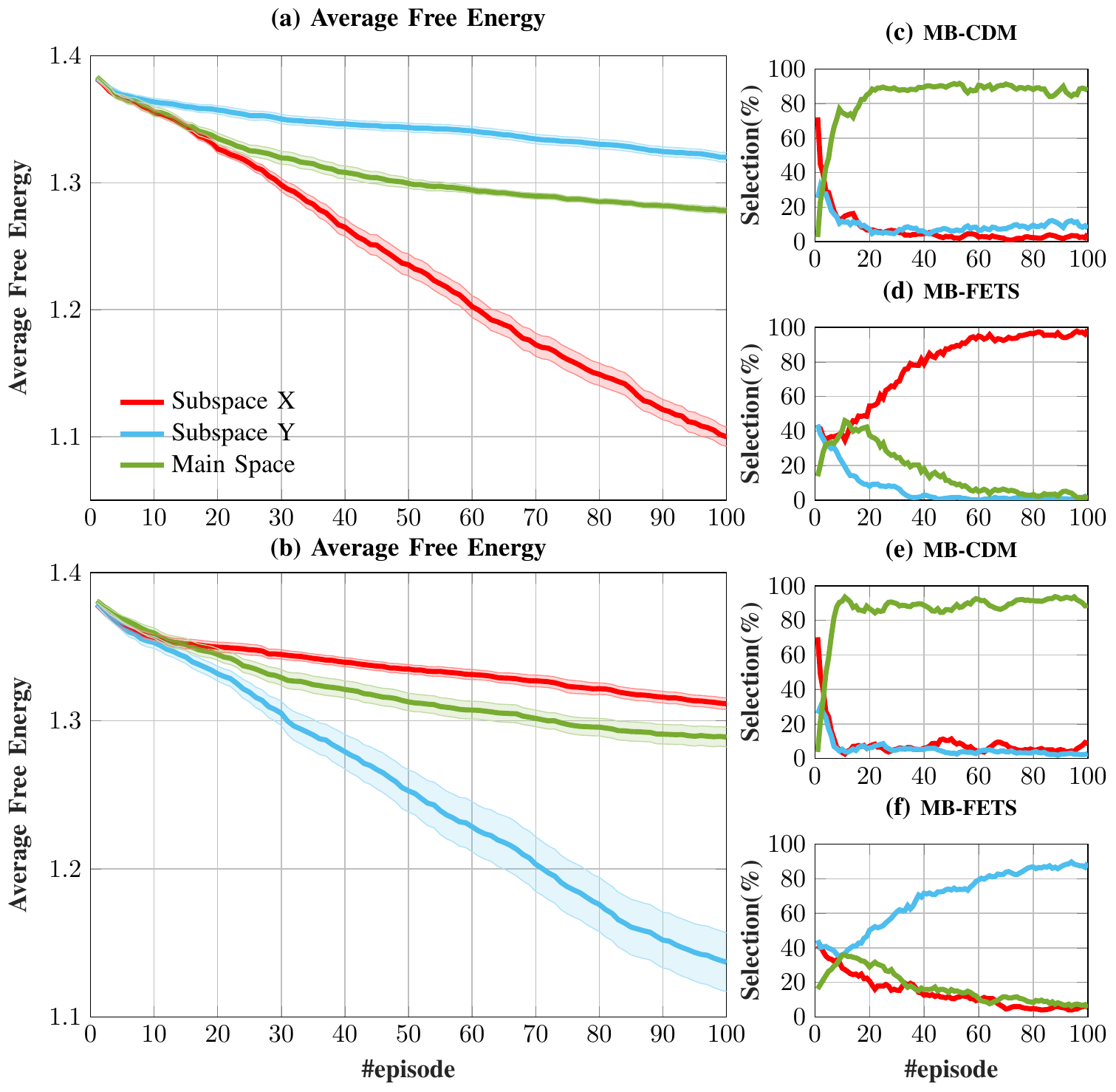}
\caption{
Average free energies and selection percentage of each model (subspaces and the main space) for model-based version of our approach in environments 7 (top plots) and 8 (bottom plots). Plots (a) and (b) are average free energy of each space as a function of the number of episodes. Plots (c)-(f) are selection percentage of each space in each episode for our and CDM approaches.}
\label{fig:FERIEnv78}
\end{figure}

Our method also has better performance in detecting which subspace is more generalizable. Environment 7 is more generalizable in subspace X (subspace that has only the X feature, and states with different Y values and the same X value are assumed to be one state.). This environment is designed such that only subspace X is informative for decision making. As it can be seen in Fig. \ref{fig:FERIEnv78}(c) and (d), our method does not stop generalization in subspace X, whereas other methods~\cite{daee2014reward,hashemzadeh2018exploiting} stop generalization by switching to the main space in the early stages of learning. Environment 8 is opposite to environment 7 and is only generalizable in subspace Y and our algorithm successfully detects this fact (see Fig. \ref{fig:FERIEnv78}(e) and (f)).

Generalizability also results in decreasing free energy. In environment 7, free energy of subspace X decreases more than subspace Y and the main space. In environment 8, it happens for subspace Y in the same way (see Fig. \ref{fig:FERIEnv78}(a) and (b)).

\subsubsection{Real-Time Strategy Game StarCraft:Broodwar}
In this subsection, we evaluate the performance of our approach in a more complex task of abstract combat maneuvering model of SC:BW between RL agent unit and enemy units\cite{wender2012applying,hashemzadeh2018exploiting}. The RL agent can move toward each enemy, retreat (get away from all enemies) and shoot each enemy and cause its hitpoint decreases. If the agent gets too close to an enemy, the enemy shoots and the agent hitpoint decreases. The agent gets the following rewards:
\begin{itemize}
\item The agent is punished by hitting the wall.
\item It is punished by firing when no enemy is nearby\footnote{The proximity to each enemy is defined by the Manhattan distance\cite{perlibakas2004distance} and if the distance is less than $\theta_E$, it means that the enemy is nearby.}.
\item It is given a reward when it kills all of the enemies.  
\item It gets a reward if it shoots an enemy and gets punishment when is shot by an enemy.
\end{itemize} 

In this environment, the agent observes location X, location Y, its hitpoint and the location and the hitpoint of each enemy. More information on the setting of SC:BW environment can be found in Appendix~\ref{apx:sc:bw}. We consider the following subspaces: the subspace of the agent XY location, the subspace of the agent's hitpoint and the subspace of the agent distance to each enemy. The subspace of the agent distance to the enemy is not primitive and is a nonlinear subspace created via a nonlinear combination of features X and Y.

\begin{figure}[!t] 
\centering
\includegraphics[width=0.5\textwidth]{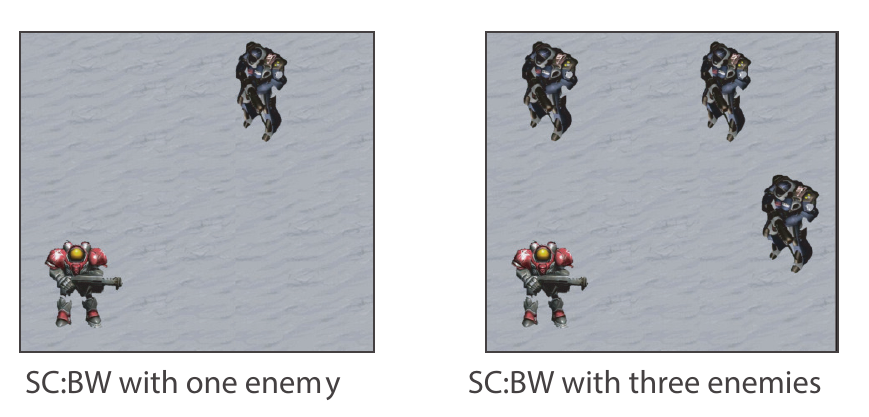}
\caption{StarCraft:BroodWar environment scenarios.}
\label{fig:1Eand3EENVs}
\end{figure}

\begin{figure}[!t] 
\centering
\includegraphics[width=0.5\textwidth]{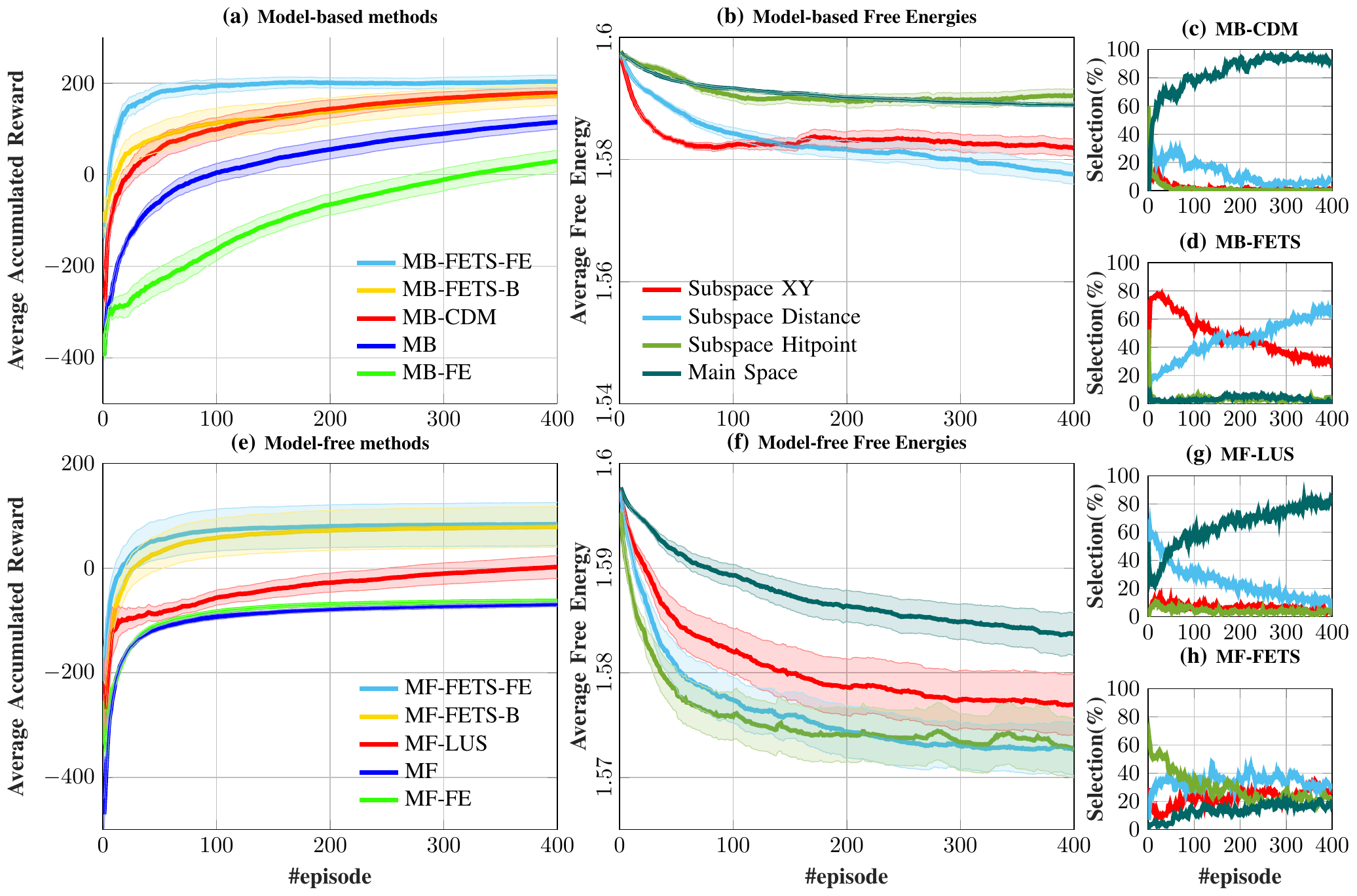}
\caption{SC:BW with one enemy results. (a) and (b) Average accumulated reward for various model-based and model-free approaches. (c) and (d) average free energy of different subspaces and main space for model-based and model-free version of our framework. (e)-(h) Selection percentage for different subspace-based methods. }
\label{fig:1EInfo}
\end{figure}

\begin{figure}[!t] 
\centering
\includegraphics[width=0.5\textwidth]{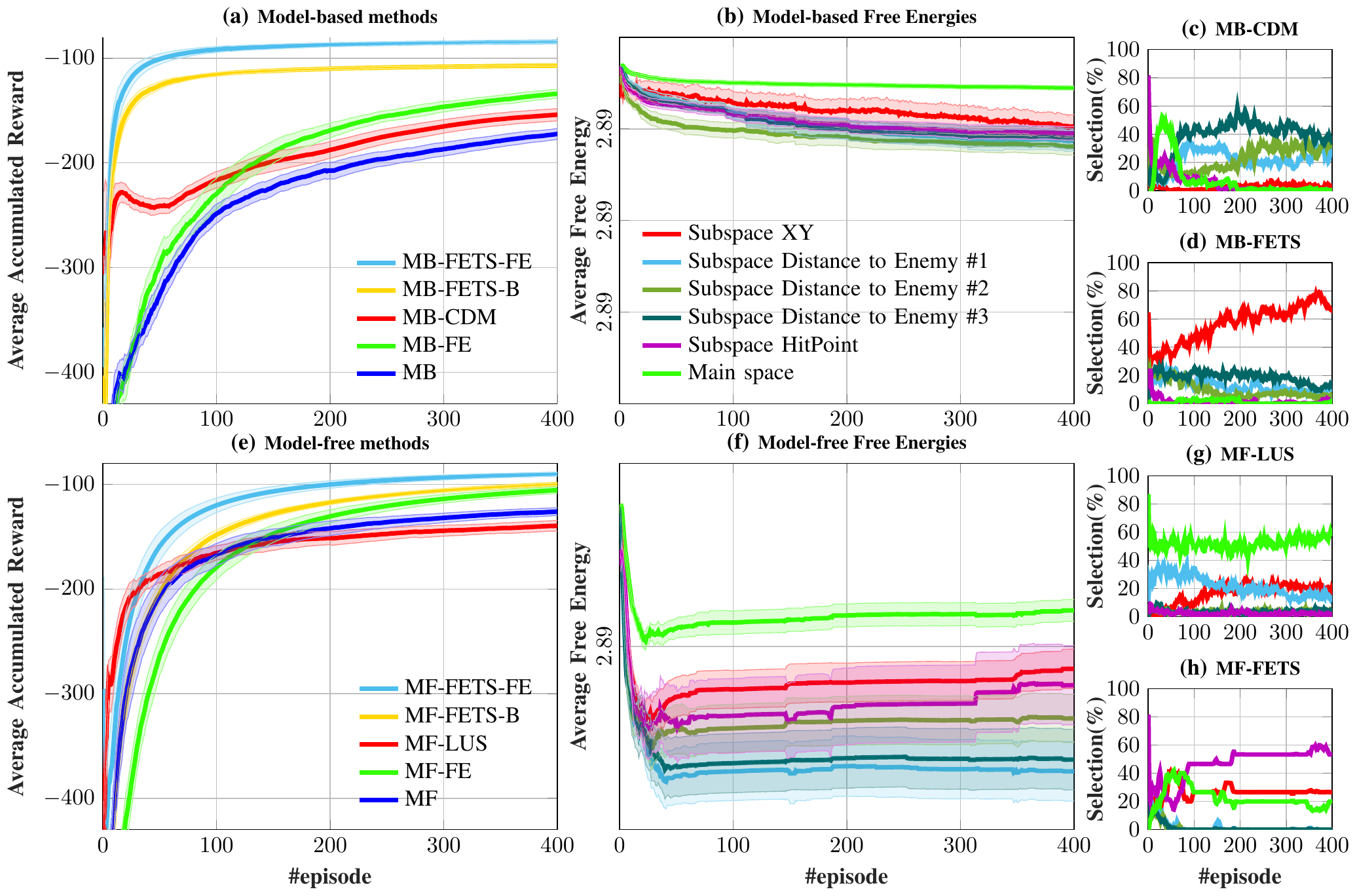}
\caption{SC:BW with three enemies results. (a) and (b) Average accumulated reward for various model-based and model-free approaches. (c) and (d) Average free energy of different subspaces and main space for the model-based and model-free version of our framework. (e)-(h) Selection percentage for different subspace-based methods.}
\label{fig:3EInfo}
\end{figure}

We simulated this environment for one enemy and three enemies scenarios depicted in Fig. \ref{fig:1Eand3EENVs}. In these scenarios, the agent starts each episode in a fixed starting state and can move, but the enemy location is fixed all over each episode. This feature of this environment makes the XY location an informative subspace.
Fig. \ref{fig:1EInfo} and Fig. \ref{fig:3EInfo} give some information about the scenarios of SC:BW environment; the average accumulated reward for different methods, average free energy and selection percentage for subspaces and main space. It can be noticed that our approach has a significantly better average accumulated reward in comparison to other methods. 

As agent gets experienced in the SC:BW environments, the subspace of distance to enemies becomes more important for the RL agent to fight with the enemies. In one enemy scenario, the agent successfully kills the enemy. As shown in Fig. \ref{fig:1EInfo}(c) and (d) free energy of the subspace of distance to enemy decreases more than other subspaces for both model-free and model-based versions of our framework in this scenario. In this environment, the confidence intervals are inacurate, and the constraint~\eqref{eq:constraint_2} helps the agent to tackle this issue.

\subsection{Continuous Domain}
\begin{figure}[!t] 
\centering
\includegraphics[width=0.5\textwidth]{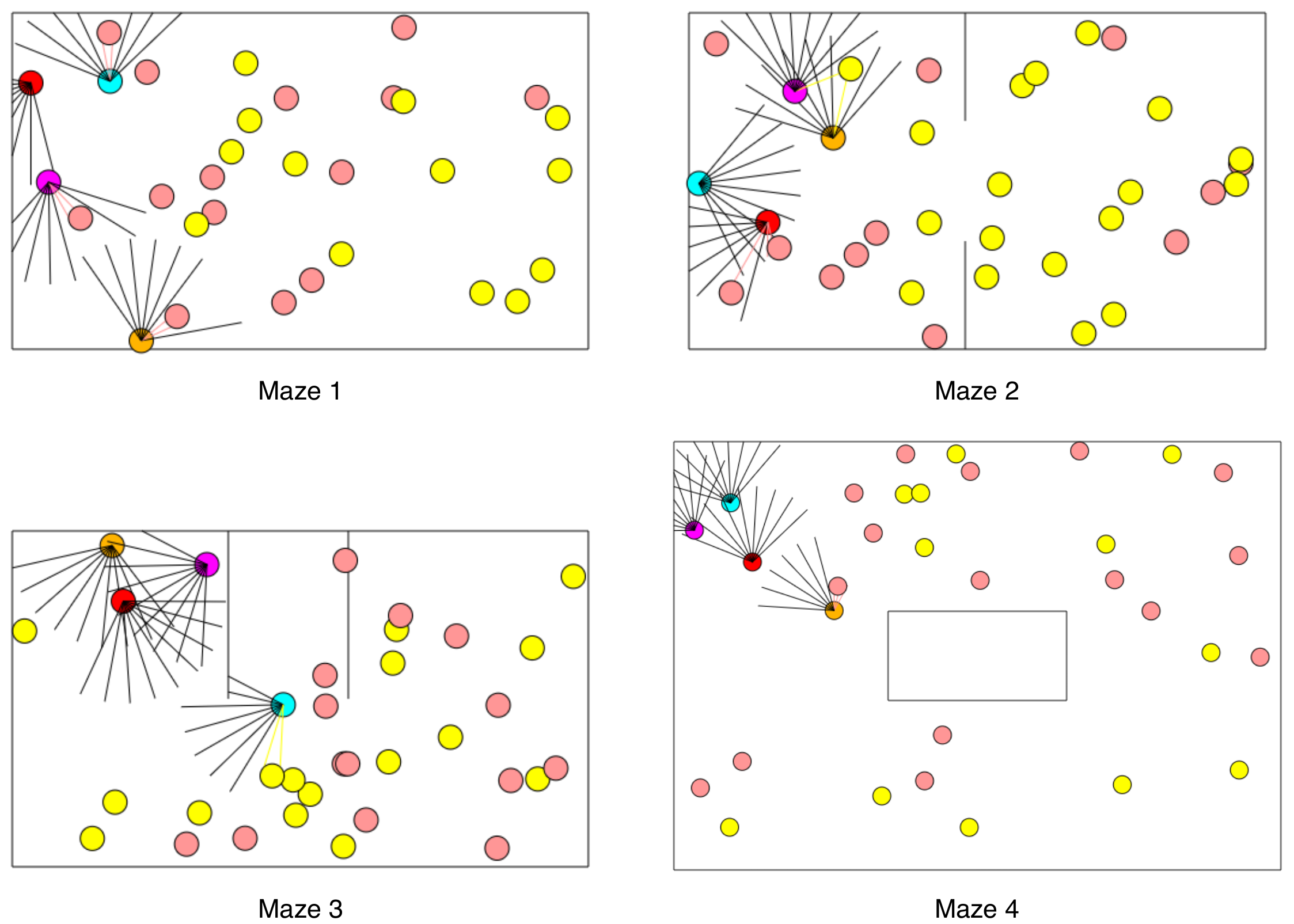}
\caption{Continuous maze environments with different barriers configurations. Different agents do not see each other and learn individually. However, they affect on a shared environment.}
\label{fig:CMEnvs}
\end{figure}

In this subsection, we also test our idea in continuous maze environments\cite{gal2016dropout}. The barrier configurations of these mazes are depicted in Fig. \ref{fig:CMEnvs}. We compare the following methods using Thompson sampling policy as the behavioral policy.
\begin{enumerate}
\item TS: Using Q-networks with dropout to get a random sample of Q-values and perform Thompson sampling policy as explained in section~\ref{ss:Dropout for Bayesian Approximation}.
\item TS-FE: The policy of the agent is given in~\eqref{eq:OptimalPolicy} instead of using the behavioral policy. 
\item TS-FETS-B: In this method, the free energy model is just used for selection between different spaces and the policy is the behavioral policy of the selected space and the sampling method is like the method of TS.
\item TS-FETS-FE: In this method, the free energy model is used both for selection between different spaces and also for integration of information of the main space into each space policy given in~\eqref{eq:OptimalPolicy}. 
\end{enumerate}
In methods that use the free energy model (all except the first), the dropout method introduced in section \ref{ss:Dropout for Bayesian Approximation} is used to estimate Thompson sampling. 
In this task, Q-networks are used for estimating Q-values.  The networks are fully-connected nets with two hidden layers. The main space Q-network has 50 neurons in each hidden layer, and the subspaces have 15 neurons, and the dropout rate is $0.1$.
This setting makes subspace Q-networks smaller, and therefore faster to learn in contrast with the main space. The free energy of subspaces decreases by reducing the number of neurons of Q-networks. The lower number of neurons also helps to have a better generalization in subspaces and also avoids over-fitting in the network.
In these environments, the parameters of the agent integration and generalization are $\alpha = \beta = 3$\footnote{We use $\alpha = \beta$ case, because in this experience the behavoiral policy is Thompson sampling and this causes simpler formalisation. The valuse are selected based on conditions $\beta \ge1$ and $\alpha>0$ of Theorem \ref{th:1}. High value of $\alpha$ can cause ignoring PA, and low value of $\alpha$ can hinder us to have good generalization. Therefore, we choose this moderate value for $\alpha$.}, respectively. In this case, since the behavioral policy is Thompson sampling and $\alpha = \beta$, we can rewrite the free energy model in~\eqref{eq:optimization_subspaces} with a simpler form:

\begin{multline}
	F(s,m,\pi(a|s,m)) = \expec_{\pi(a|s,m)}\left[-\log{\pi_{TS}(a|s,m)}\right] + \\
	\frac{1}{\alpha} \mathbb{D}_{KL}( \pi(a|s,m) || \pi_{TS}(a|s,m_{Main})) .
\end{multline}

This means that the agent is maximizing expected utility defined in~\eqref{eq:utility} having the following constraint that the produced policy by free energy optimization stays close to the policy of the main space as follows

\begin{equation}
    \mathbb{D}_{KL}( \pi(a|s,m) || \pi_{TS}(a|s,m_{Main})) <K,
\end{equation}

In this environment, the agent has nine eyes, pointing to different directions. Each eye can observe three features that are proximity to wall, poison or food so that 27 sensors can determine the environment states. The agent can move toward five directions and gets a reward by colliding to red circles (food) and gets punished by colliding to yellow circles (poison). The agent gets further reward for not going toward walls and move straight in a line\cite{gal2016dropout}. The task which is not stationary, and the red circles and green ones appear in random locations after the agent eats them, so the location information is redundant in this task unlike the SC:BW environment. Information on the setting of this environment can be found in Appendix~\ref{apx:cont_env}.

\begin{figure}[!t] 
\centering
\includegraphics[width=0.5\textwidth]{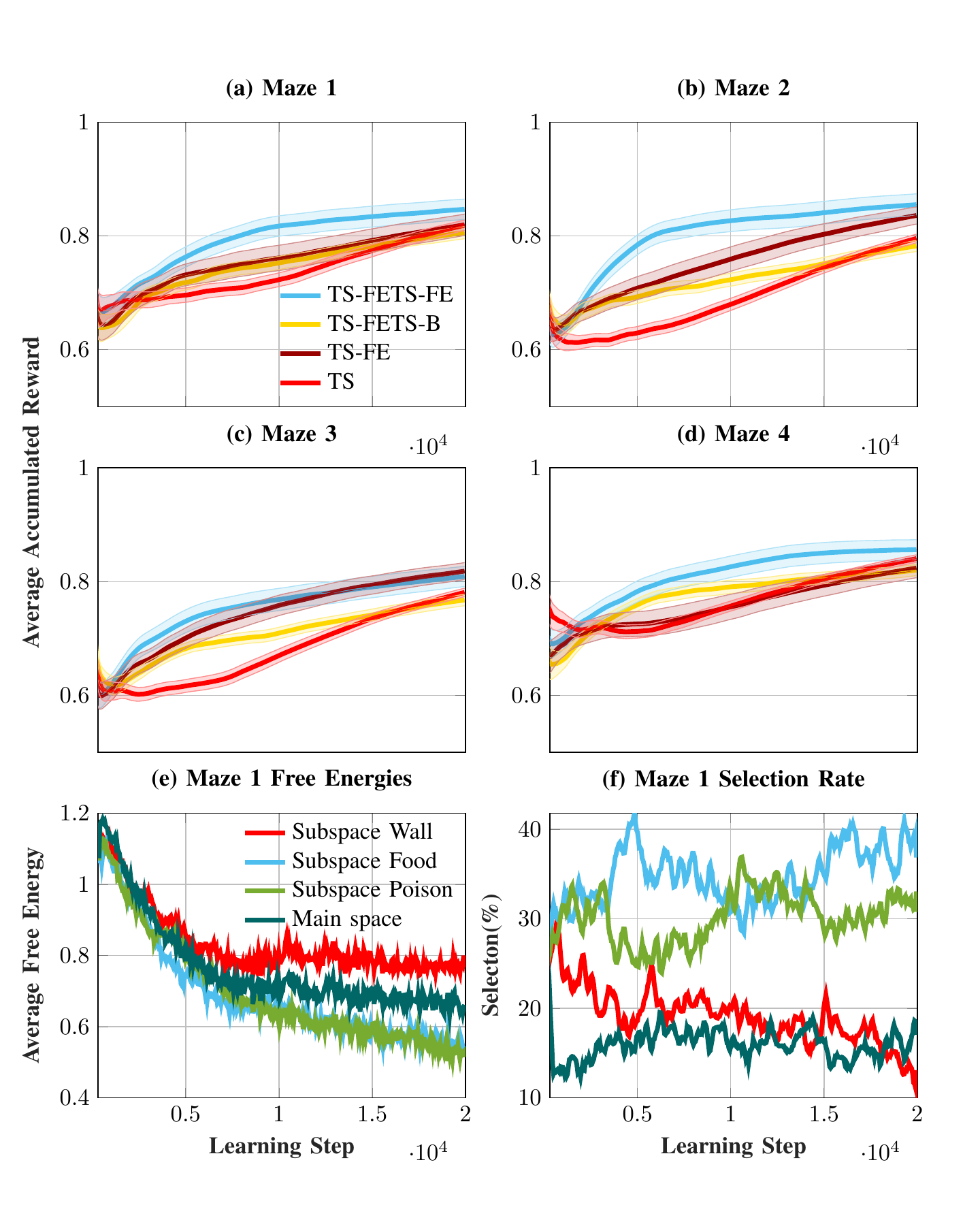}
\caption{Continuous environments results. (a)-(d) Average accumulated reward of different methods in different environments. (e) Average free energy for each subspace in maze 1. (f) Selection percentage of each subspace in maze 1. }
\label{fig:CMInfo}
\end{figure}

For an accurate comparison of different approaches, we perform a multi-agent simulation similar to~\cite{gal2016dropout} where each agent strategy corresponds to one of four alternative methods.
Fig. \ref{fig:CMInfo} depicts average accumulated reward, average free energies and selection percentage of each subspace. Our approach improves Thompson sampling in early trials in the learning process.

As the agent gets experienced in the environment, it avoids eating poisons and moves toward foods. Free energies of the subspaces of food and poison decrease more than the subspace of wall (Fig. \ref{fig:CMInfo}(e) and (f)).

\section{Conclusion}
In this paper, we introduced a framework to improve sample efficiency of reinforcement learning algorithms using the generalization property of subspaces but in face of higher computational complexity in both discrete and continuous domains. The computational complexity is a direct effect of learning in subspaces alongside the main space. Subspaces help the agent sample efficiency because they have a fewer number of parameters, so they can learn the task faster than the main space, but in general, they cannot reach the optimal policy. Therefore, we proposed the free energy model for space selection and also policy generation.

The free energy model is suitable to model decision-making under some resource limitations. In this work, the resource limitations are sub-optimality of the suggested policies by the subspaces and also inaccurate uncertainty estimations. Therefore, to tackle them, we used the free energy model, where we defined utility to be the optimality of actions in a state and considered information constraints to guarantee the generated policy remains consistent with information of the main space and also is close enough to a behavioral policy. 

We used the optimal free energy as a criterion to select between different spaces. Then, we considered two cases for policy: the behavioral policy, and the policy suggested by the free energy model. We saw that in most problems the policy that is suggested by the free energy model has a better performance. This policy in each space helps to avoid over-exploration which is one of the limitations of using Thompson sampling in large-scale environments.

Our approach is robust to inaccurate uncertainty estimation (for example in SC:BW), and this feature makes it also usable in model-free approaches where the confidence intervals are less accurate than model-based approaches.
Also, we observed that our method can confront the PA problem of the subspaces better than the standard RL approach and also previous related methods. This was evident from the high performance of our method in comparison with other approaches.

We saw that in our framework non-informative subspaces get higher free energy as the agent gets more experienced (see Fig. \ref{fig:FERIEnv78} and Fig. \ref{fig:CMInfo}). If the subspaces are created by selecting a subset of features, removing redundant features from the main space can reduce the computational complexity of the algorithm. This scheme related to the feature-based attention control in the literature~\cite{borji2009learning,borji2010online}. Another usage of detecting non-informative subspaces is to remove them during the learning process to reduce the computational complexity of the algorithm. and our method can detect this possibility.

There are multiple directions for future works:
\begin{itemize}
\item Finding an optimal set of subspaces which is task-based can be done automatically using a criterion. If each subspace corresponds a sub-task, then more complex problems like SC:BW can be solved.
\item Although our method has robustness to inaccurate uncertainty estimation, more reliable approaches to estimate Thompson sampling policy can improve the sample efficiency.
\item In the current method, the temperature of free energy model is not automatically determined. An adaptive method to set the temperature might cause improvements.
\item In continuous domains, automatically adapting of network settings like number of neurons, learning rate, and etc. can improve the robustness of the method. Alternatively, we can have multiple networks with different hyperparameters and treat them as different subspaces.

\end{itemize}

\nocite{*}
\bibliographystyle{./sty/IEEETran}
\bibliography{./ref/myref}

\begin{thebibliography}{10}
\providecommand{\url}[1]{#1}
\csname url@samestyle\endcsname
\providecommand{\newblock}{\relax}
\providecommand{\bibinfo}[2]{#2}
\providecommand{\BIBentrySTDinterwordspacing}{\spaceskip=0pt\relax}
\providecommand{\BIBentryALTinterwordstretchfactor}{4}
\providecommand{\BIBentryALTinterwordspacing}{\spaceskip=\fontdimen2\font plus
\BIBentryALTinterwordstretchfactor\fontdimen3\font minus
  \fontdimen4\font\relax}
\providecommand{\BIBforeignlanguage}[2]{{%
\expandafter\ifx\csname l@#1\endcsname\relax
\typeout{** WARNING: IEEEtran.bst: No hyphenation pattern has been}%
\typeout{** loaded for the language `#1'. Using the pattern for}%
\typeout{** the default language instead.}%
\else
\language=\csname l@#1\endcsname
\fi
#2}}
\providecommand{\BIBdecl}{\relax}
\BIBdecl

\bibitem{kaelbling1996reinforcement}
L.~P. Kaelbling, M.~L. Littman, and A.~W. Moore, ``Reinforcement learning: A
  survey,'' \emph{Journal of artificial intelligence research}, vol.~4, pp.
  237--285, 1996.

\bibitem{abbeel2006learning}
P.~Abbeel, V.~Ganapathi, and A.~Y. Ng, ``Learning vehicular dynamics, with
  application to modeling helicopters,'' in \emph{Advances in Neural
  Information Processing Systems}, 2006, pp. 1--8.

\bibitem{navarro2012real}
N.~Navarro-Guerrero, C.~Weber, P.~Schroeter, and S.~Wermter, ``Real-world
  reinforcement learning for autonomous humanoid robot docking,''
  \emph{Robotics and Autonomous Systems}, vol.~60, no.~11, pp. 1400--1407,
  2012.

\bibitem{imanberdiyev2016autonomous}
N.~Imanberdiyev, C.~Fu, E.~Kayacan, and I.-M. Chen, ``Autonomous navigation of
  uav by using real-time model-based reinforcement learning,'' in \emph{2016
  14th International Conference on Control, Automation, Robotics and Vision
  (ICARCV)}.\hskip 1em plus 0.5em minus 0.4em\relax IEEE, 2016, pp. 1--6.

\bibitem{lu2015recommender}
J.~Lu, D.~Wu, M.~Mao, W.~Wang, and G.~Zhang, ``Recommender system application
  developments: a survey,'' \emph{Decision Support Systems}, vol.~74, pp.
  12--32, 2015.

\bibitem{da2016evolutionary}
E.~Q. da~Silva, C.~G. Camilo-Junior, L.~M.~L. Pascoal, and T.~C. Rosa, ``An
  evolutionary approach for combining results of recommender systems techniques
  based on collaborative filtering,'' \emph{Expert Systems with Applications},
  vol.~53, pp. 204--218, 2016.

\bibitem{holzinger2016interactive}
A.~Holzinger, ``Interactive machine learning for health informatics: when do we
  need the human-in-the-loop?'' \emph{Brain Informatics}, vol.~3, no.~2, pp.
  119--131, 2016.

\bibitem{zhang2017human}
J.~Zhang, P.~Fiers, K.~A. Witte, R.~W. Jackson, K.~L. Poggensee, C.~G. Atkeson,
  and S.~H. Collins, ``Human-in-the-loop optimization of exoskeleton assistance
  during walking,'' \emph{Science}, vol. 356, no. 6344, pp. 1280--1284, 2017.

\bibitem{li2016dialogue}
J.~Li, A.~H. Miller, S.~Chopra, M.~Ranzato, and J.~Weston, ``Dialogue learning
  with human-in-the-loop,'' \emph{arXiv preprint arXiv:1611.09823}, 2016.

\bibitem{sutton1996generalization}
R.~S. Sutton, ``Generalization in reinforcement learning: Successful examples
  using sparse coarse coding,'' in \emph{Advances in neural information
  processing systems}, 1996, pp. 1038--1044.

\bibitem{thompson1933likelihood}
W.~R. Thompson, ``On the likelihood that one unknown probability exceeds
  another in view of the evidence of two samples,'' \emph{Biometrika}, vol.~25,
  no. 3/4, pp. 285--294, 1933.

\bibitem{gopalan2014thompson}
A.~Gopalan, S.~Mannor, and Y.~Mansour, ``Thompson sampling for complex online
  problems,'' in \emph{International Conference on Machine Learning}, 2014, pp.
  100--108.

\bibitem{mandel2016efficient}
T.~Mandel, Y.-E. Liu, E.~Brunskill, and Z.~Popovic, ``Efficient bayesian
  clustering for reinforcement learning.'' in \emph{IJCAI}, 2016, pp.
  1830--1838.

\bibitem{abel2017near}
D.~Abel, D.~E. Hershkowitz, and M.~L. Littman, ``Near optimal behavior via
  approximate state abstraction,'' \emph{arXiv preprint arXiv:1701.04113},
  2017.

\bibitem{abel2018state}
D.~Abel, D.~Arumugam, L.~Lehnert, and M.~Littman, ``State abstractions for
  lifelong reinforcement learning,'' in \emph{International Conference on
  Machine Learning}, 2018, pp. 10--19.

\bibitem{firouzi2011interactive}
H.~Firouzi, M.~N. Ahmadabadi, B.~N. Araabi, S.~Amizadeh, M.~S. Mirian, and
  R.~Siegwart, ``Interactive learning in continuous multimodal space: A
  bayesian approach to action-based soft partitioning and learning,''
  \emph{IEEE Transactions on Autonomous Mental Development}, vol.~4, no.~2, pp.
  124--138, 2011.

\bibitem{daee2014reward}
P.~Daee, M.~S. Mirian, and M.~N. Ahmadabadi, ``Reward maximization justifies
  the transition from sensory selection at childhood to sensory integration at
  adulthood,'' \emph{PloS one}, vol.~9, no.~7, p. e103143, 2014.

\bibitem{hashemzadeh2018exploiting}
M.~Hashemzadeh, R.~Hosseini, and M.~N. Ahmadabadi, ``Exploiting generalization
  in the subspaces for faster model-based reinforcement learning,'' \emph{IEEE
  transactions on neural networks and learning systems}, vol.~30, no.~6, pp.
  1635--1650, 2018.

\bibitem{silva2018object}
F.~L.~D. Silva and A.~H.~R. Costa, ``Object-oriented curriculum generation for
  reinforcement learning,'' in \emph{Proceedings of the 17th International
  Conference on Autonomous Agents and MultiAgent Systems}.\hskip 1em plus 0.5em
  minus 0.4em\relax International Foundation for Autonomous Agents and
  Multiagent Systems, 2018, pp. 1026--1034.

\bibitem{schmidhuber2013powerplay}
J.~Schmidhuber, ``Powerplay: Training an increasingly general problem solver by
  continually searching for the simplest still unsolvable problem,''
  \emph{Frontiers in psychology}, vol.~4, p. 313, 2013.

\bibitem{narvekar2019learning}
S.~Narvekar and P.~Stone, ``Learning curriculum policies for reinforcement
  learning,'' in \emph{Proceedings of the 18th International Conference on
  Autonomous Agents and MultiAgent Systems}.\hskip 1em plus 0.5em minus
  0.4em\relax International Foundation for Autonomous Agents and Multiagent
  Systems, 2019, pp. 25--33.

\bibitem{narvekar2020curriculum}
S.~Narvekar, B.~Peng, M.~Leonetti, J.~Sinapov, M.~E. Taylor, and P.~Stone,
  ``Curriculum learning for reinforcement learning domains: A framework and
  survey,'' \emph{arXiv preprint arXiv:2003.04960}, 2020.

\bibitem{tesauro1995temporal}
G.~Tesauro, ``Temporal difference learning and td-gammon,''
  \emph{Communications of the ACM}, vol.~38, no.~3, pp. 58--68, 1995.

\bibitem{silver2016mastering}
D.~Silver, A.~Huang, C.~J. Maddison, A.~Guez, L.~Sifre, G.~Van Den~Driessche,
  J.~Schrittwieser, I.~Antonoglou, V.~Panneershelvam, M.~Lanctot \emph{et~al.},
  ``Mastering the game of go with deep neural networks and tree search,''
  \emph{nature}, vol. 529, no. 7587, p. 484, 2016.

\bibitem{vezhnevets2016strategic}
A.~Vezhnevets, V.~Mnih, S.~Osindero, A.~Graves, O.~Vinyals, J.~Agapiou
  \emph{et~al.}, ``Strategic attentive writer for learning macro-actions,'' in
  \emph{Advances in neural information processing systems}, 2016, pp.
  3486--3494.

\bibitem{ortega2011information}
D.~A. Ortega and P.~A. Braun, ``Information, utility and bounded rationality,''
  in \emph{International Conference on Artificial General Intelligence}.\hskip
  1em plus 0.5em minus 0.4em\relax Springer, 2011, pp. 269--274.

\bibitem{ortega2013thermodynamics}
P.~A. Ortega and D.~A. Braun, ``Thermodynamics as a theory of decision-making
  with information-processing costs,'' \emph{Proceedings of the Royal Society
  A: Mathematical, Physical and Engineering Sciences}, vol. 469, no. 2153, p.
  20120683, 2013.

\bibitem{stephan2009bayesian}
K.~E. Stephan, W.~D. Penny, J.~Daunizeau, R.~J. Moran, and K.~J. Friston,
  ``Bayesian model selection for group studies,'' \emph{Neuroimage}, vol.~46,
  no.~4, pp. 1004--1017, 2009.

\bibitem{bellman1957markovian}
R.~Bellman, ``A markovian decision process,'' \emph{Journal of mathematics and
  mechanics}, pp. 679--684, 1957.

\bibitem{sutton2018reinforcement}
R.~S. Sutton and A.~G. Barto, \emph{Reinforcement learning: An
  introduction}.\hskip 1em plus 0.5em minus 0.4em\relax MIT press, 2018.

\bibitem{ortega2014generalized}
P.~A. Ortega and D.~A. Braun, ``Generalized thompson sampling for sequential
  decision-making and causal inference,'' \emph{Complex Adaptive Systems
  Modeling}, vol.~2, no.~1, p.~2, 2014.

\bibitem{gal2016dropout}
Y.~Gal and Z.~Ghahramani, ``Dropout as a bayesian approximation: Representing
  model uncertainty in deep learning,'' in \emph{international conference on
  machine learning}, 2016, pp. 1050--1059.

\bibitem{jaksch2010near}
T.~Jaksch, R.~Ortner, and P.~Auer, ``Near-optimal regret bounds for
  reinforcement learning,'' \emph{Journal of Machine Learning Research},
  vol.~11, no. Apr, pp. 1563--1600, 2010.

\bibitem{daee2014developmental}
P.~Daee, ``A developmental method for multimodal sensory integration,'' Ph.D.
  dissertation, MS thesis, School Elect. Comput. Eng., Univ. Tehran, Tehran,
  Iran, 2014.

\bibitem{hoeffding2014probability}
W.~Hoeffding, ``Probability inequalities for sums of bounded random
  variables,'' \emph{Wiley StatsRef: Statistics Reference Online}, 2014.

\bibitem{weissman2003inequalities}
T.~Weissman, E.~Ordentlich, G.~Seroussi, S.~Verdu, and M.~J. Weinberger,
  ``Inequalities for the l1 deviation of the empirical distribution,''
  \emph{Hewlett-Packard Labs, Tech. Rep}, 2003.

\bibitem{white2010interval}
M.~White and A.~White, ``Interval estimation for reinforcement-learning
  algorithms in continuous-state domains,'' in \emph{Advances in Neural
  Information Processing Systems}, 2010, pp. 2433--2441.

\bibitem{tishby2011information}
N.~Tishby and D.~Polani, ``Information theory of decisions and actions,'' in
  \emph{Perception-action cycle}.\hskip 1em plus 0.5em minus 0.4em\relax
  Springer, 2011, pp. 601--636.

\bibitem{gottwald2019bounded}
S.~Gottwald and D.~A. Braun, ``Bounded rational decision-making from elementary
  computations that reduce uncertainty,'' \emph{Entropy}, vol.~21, no.~4, p.
  375, 2019.

\bibitem{trujillo2019mental}
L.~T. Trujillo, ``Mental effort and information-processing costs are
  inversely-related to global brain free energy during visual categorization,''
  \emph{Frontiers in neuroscience}, vol.~13, p. 1292, 2019.

\bibitem{hihn2019hierarchical}
H.~Hihn and D.~A. Braun, ``Hierarchical expert networks for meta-learning,''
  \emph{arXiv preprint arXiv:1911.00348}, 2019.

\bibitem{hihn2019information}
H.~Hihn, S.~Gottwald, and D.~A. Braun, ``An information-theoretic on-line
  learning principle for specialization in hierarchical decision-making
  systems,'' in \emph{2019 IEEE 58th Conference on Decision and Control
  (CDC)}.\hskip 1em plus 0.5em minus 0.4em\relax IEEE, 2019, pp. 3677--3684.

\bibitem{grau2016planning}
J.~Grau-Moya, F.~Leibfried, T.~Genewein, and D.~A. Braun, ``Planning with
  information-processing constraints and model uncertainty in markov decision
  processes,'' in \emph{Joint European Conference on Machine Learning and
  Knowledge Discovery in Databases}.\hskip 1em plus 0.5em minus 0.4em\relax
  Springer, 2016, pp. 475--491.

\bibitem{watkins1989learning}
C.~Watkins, ``Learning from delayed rewards (phd dissertation),''
  \emph{King’s College Cambridge, England}, 1989.

\bibitem{wender2012applying}
S.~Wender and I.~Watson, ``Applying reinforcement learning to small scale
  combat in the real-time strategy game starcraft: Broodwar,'' in \emph{2012
  IEEE Conference on Computational Intelligence and Games (CIG)}.\hskip 1em
  plus 0.5em minus 0.4em\relax IEEE, 2012, pp. 402--408.

\bibitem{perlibakas2004distance}
V.~Perlibakas, ``Distance measures for pca-based face recognition,''
  \emph{Pattern recognition letters}, vol.~25, no.~6, pp. 711--724, 2004.

\bibitem{borji2009learning}
A.~Borji, M.~N. Ahmadabadi, and B.~N. Araabi, ``Learning sequential visual
  attention control through dynamic state space discretization,'' in \emph{2009
  IEEE International Conference on Robotics and Automation}.\hskip 1em plus
  0.5em minus 0.4em\relax IEEE, 2009, pp. 2258--2263.

\bibitem{borji2010online}
A.~Borji, M.~N. Ahmadabadi, B.~N. Araabi, and M.~Hamidi, ``Online learning of
  task-driven object-based visual attention control,'' \emph{Image and Vision
  Computing}, vol.~28, no.~7, pp. 1130--1145, 2010.

\bibitem{callen1998thermodynamics}
H.~B. Callen, ``Thermodynamics and an introduction to thermostatistics,'' 1998.

\end{thebibliography}
\appendices

\section{The Proof of Theorem \ref{th:1}}\label{apx:proof_th1}

The optimal free energy is related to $Z(s,m)$ as follows~\cite{callen1998thermodynamics} 
	 \begin{equation*}
	 F^*(s,m) = -\frac{1}{\alpha}\log{Z(s,m)},
	 \end{equation*}
	 \begin{equation*}
	 Z(s,m) = \sum_a \pi_{B}(a|s,m) e^{\alpha \tilde{U}(a,s,m)}.
	 \end{equation*}
	 
	 Since all the logarithms are bounded by adding a small constant $\xi>0$ to the Thompson sampling probabilities, and the behavioral policy is $\epsilon$-greedy and $\alpha$ is finite, we have
	\begin{equation*}
	\pi_{B}(a|s,m_{Main}) e^{\alpha \tilde{U}(a,s,m)}\ge \frac{\epsilon}{|A|} e^{\alpha \log{\xi}}=\frac{\epsilon}{|A|} \xi^\alpha >0.
	 \end{equation*}
	 Since $\tilde{U}(a,s,m)\le 0$, we have
	 \begin{equation*}
	 Z(s,m) = \sum_a \pi_{B}(a|s,m) e^{\alpha \tilde{U}(a,s,m)}\le\sum_a \pi_{B}(a|s,m)=1,
	 \end{equation*}
	 and therefore
	 \begin{equation*}
	\pi^*(a|s,m) = \frac{1}{Z(s,m)}\pi_{B}(a|s,m_{Main}) e^{\alpha \tilde{U}(a,s,m)}>0,
\end{equation*}
which shows that the agent does not stop exploration. When the sample size $n(s,a)$ goes to infinity, and the confidence intervals go to zero, the Thompson sampling policies become one-hot, that is for one action it is equal to 1 and for the others it is equal to $\xi$. Without losing the generality, we suppose that the first action is optimal in the main space, and the second one is optimal in the subspace $m$, so $\tilde{U}(.,s,m)$\footnote{$\tilde{U}(.,s,m)$ is a vector with elements $\tilde{U}(a,s,m)$ for different values of $a$. } and $\tilde{U}(.,s,m_{Main})$ are computed as follows:
	 \begin{multline*}
	 \tilde{U}(.,s,m) = \frac{1}{\beta}\log{\pi_{TS}(.|s,m_{Main})} +\\
	 \frac{\beta-1}{\beta}\log{\pi_{TS}(.|s,m)}=\\
	 \left[\frac{\beta-1}{\beta}\log(\xi) , \frac{1}{\beta}\log(\xi) , \log(\xi) , \cdots , \log(\xi)\right],
	 \end{multline*}
	 \begin{multline*}
	 \tilde{U}(.,s,m_{Main}) = \frac{1}{\beta}\log{\pi_{TS}(.|s,m_{Main})} +\\
	 \frac{\beta-1}{\beta}\log{\pi_{TS}(.|s,m_{Main})}=\\
	 \left[0 , \log(\xi), \cdots , \log(\xi)\right],
	 \end{multline*}
	 and the behavioral policy for them is $\epsilon$-greedy and is written as follows
	 \begin{equation*}
	 \pi_B(.|s,m) = \left[\frac{\epsilon}{|A|} , 1-\epsilon+\frac{\epsilon}{|A|} , \frac{\epsilon}{|A|} , \cdots,\frac{\epsilon}{|A|}\right],
	 \end{equation*}
	 \begin{equation*}
	 \pi_B(.|s,m_{Main}) = \left[1-\epsilon+\frac{\epsilon}{|A|} , \frac{\epsilon}{|A|} , \cdots,\frac{\epsilon}{|A|}\right],
	 \end{equation*}
	 so we have
	 \begin{multline*}
	 Z(s,m) = \frac{\epsilon}{|A|} \xi^{\frac{\alpha(\beta-1)}{\beta}} + (1-\epsilon+\frac{\epsilon}{|A|})\xi^{\frac{\alpha}{\beta}}\\
	 +\frac{(|A|-2)\epsilon}{|A|}\xi^{\alpha },
	 \end{multline*}
	 \begin{equation*}
	 Z(s,m_{Main}) = (1-\epsilon+\frac{\epsilon}{|A|})e^{0}+\frac{(|A|-1)\epsilon}{|A|}\xi^{\alpha},
	 \end{equation*}
	 and therefore
	 \begin{multline*}
	 Z(s,m)-Z(s,m_{Main}) = \frac{\epsilon}{|A|}\big(\xi^{\frac{\beta-1}{\beta}\alpha}+\xi^{\frac{\alpha}{\beta}}-\xi^{\alpha}-1\big)\\
	 +(1-\epsilon)(-1+\xi^{\frac{\alpha}{\beta}}).
	 \end{multline*}

	 Since $\alpha>0$ and $\beta>1$, we can find $\xi>0$ such that 
	 \begin{equation*}
	 \frac{\epsilon}{|A|}\big(\xi^{\frac{\beta-1}{\beta}\alpha}+\xi^{\frac{\alpha}{\beta}}-\xi^{\alpha}\big)
	 +(1-\epsilon)(\xi^{\frac{\alpha}{\beta}})<\frac{\epsilon}{|A|}.
	 \end{equation*}
	 
	 So we have
	 \begin{equation*}
	 Z(s,m) - Z(s,m_{Main})<\epsilon-1<0 ,
	 \end{equation*}
	 and therefore
	 \begin{equation*}
	 F^*(s,m_{Main})<F^*(s,m),
	 \end{equation*}
	 so if the optimal policy for a subspace differs from the main space, the algorithm selects the main space. Since the agent finally uses the main space and it explores all actions in all states infinitely, it finally finds the true value for all state-actions and converges to the optimal policy.

\section{Settings of 2-D Maze Environment}\label{apx:2d_maze}
In this environment, the state of the agent is its 2D position and it changes its position using four actions. If there is a barrier in the new position, the agent remains in the current location and gets reward RW. The agent obtains reward RS for taking each action. If the agent reaches the goal, it gets reward RG~\cite{hashemzadeh2018exploiting}. The distribution of these immediate rewards are shown in Table~\ref{T:Envparameter}.
\begin{table} 
  \small
  \begin{center}
    \caption{Distribution of the rewards for different scenarios in the 2D maze task~\cite{hashemzadeh2018exploiting}.}
    \label{T:Envparameter}
    \begin{tabular}{|c|c|}%
	    \hline		
	    \textbf{Symbol} 	&	\textbf{Distribution}	
	    \\		\hline  \hline		
	    RW  	&	$\sim \frac{1}{3} \mathcal{N}(-11.5, 0.2) + \frac{2}{3} \mathcal{N}(-10.5, 0.3) \; \in [-12, -10]$ 	
		    \\	 \hline		
	    RG	&	$\sim N(+10, 0.02) \; \in [9.5, 11.5]$	
	    \\		 \hline
	     RS 	&	$\sim \frac{1}{3} \mathcal{N}(-1.5, 0.2) + \frac{2}{3} \mathcal{N}(-0.5, 0.3) \; \in [-2, 0]$	
	    \\		\hline		
      \end{tabular} 
  \end{center}
\end{table}
\section{Settings of Real-Time Strategy Game StarCraft:Broodwar}\label{apx:sc:bw}
In this environment, state of the agent is its 2D position and its hitpoint and the enemies hitpoints. It has the folowing abstract actions: moving toward an enemy, retreating to the starting state, and shooting to each enemy. The agent's goal is to eliminate all enemies. The reward for doing action $a$ from state $s$ and going to state $s'$ is computed by the following relation~\cite{hashemzadeh2018exploiting}:
\begin{equation}
r = 10\times \left( \sum_{i=1}^{\#\text{enemies}} \left( HP_i - HP'_i \right) - \left(HP - HP'\right) \right) +r_{a} - 1  ,
\end{equation}
where $HP_i$ and $HP'_i$ are the hitpoints of the ith enemy, where the agent is in states $s$ and $s'$, respectively. $HP$ and $HP'$ denote the agent hitpoints in $s$ and $s'$, respectively. $r_a$ is the extra reward that is given to the agent when it shoots a killed enemy (RA), when it hits to the wall (RW) and when it attains the goal of killing all enemies (RG). The distribution of $r_a$ is shown in Table~\ref{T:SCRewrd}.
\begin{table}
	\small
	\begin{center}
		\caption{Distribution of the rewards for SC:BW task~\cite{hashemzadeh2018exploiting}.}
		\label{T:SCRewrd}
		\begin{tabular}{|c|c|}%
			\hline		
			\textbf{Symbol}	& \textbf{Distribution}
			\\		\hline  \hline		
			RA	&	$\sim N(-10, 0.1), \; \in [-11, -9]$ 	
			\\	 \hline		
			RW	& $\sim \frac{1}{3} \mathcal{N}(-1.5, 0.2) + \frac{2}{3} \mathcal{N}(-0.5, 0.3), \; \in [-2, 0]$	
			\\		 \hline
			RG&	$\sim \frac{1}{3} \mathcal{N}(97, 0.2) + \frac{2}{3} \mathcal{N}(100, 0.15), \; \in [95, 105]$	
			\\		\hline		
		\end{tabular} 
	\end{center}
\end{table}
\section{Settings of Countinuous Maze Environment}\label{apx:cont_env}
In this environment, the state of the agent is the proximity of its nine eyes to walls, foods, and poisons. Its actions are moving toward five different directions in the continuous maze. If it collides with poisons or foods, it digests them. The reward of doing action $a$ is computed as follows
\begin{equation}
r = R_{P} + R_{SF} + R_{D} ,
\end{equation}
where $R_{P}$ is the average of proximity to a wall for different eyes and the distance is normalized by dividing by the maximum possible proximity. $R_{SF}$ is the straight forward reward to make the agent walk in a straight line. If the agent takes the forward action and $R_P>0.75$, $R_{SF} = 0.1 R_{P}$, otherwise $R_{SF} = 0$. $R_D$ is the digestion reward. If the agent eats a food, $R_D = 5$, and if it eats a poison, $R_D = -6$.

\end{document}